\documentclass{article}

\usepackage[eandd,final]{neurips_2026}

\usepackage[utf8]{inputenc}
\usepackage[T1]{fontenc}
\usepackage{url}
\makeatletter
\g@addto@macro\UrlBreaks{\do\-\do\.}
\makeatother
\usepackage{hyperref}
\usepackage{booktabs}
\usepackage{amsfonts}
\usepackage{nicefrac}
\usepackage{microtype}
\usepackage{xcolor}

\usepackage{wrapfig}
\usepackage{graphicx}

\usepackage{amsthm}

\usepackage{amsmath,amsfonts,bm}

\def\eqref#1{equation~\ref{#1}}

\def\1{\bm{1}}

\DeclareMathAlphabet{\mathsfit}{\encodingdefault}{\sfdefault}{m}{sl}
\SetMathAlphabet{\mathsfit}{bold}{\encodingdefault}{\sfdefault}{bx}{n}

\theoremstyle{plain}
\newtheorem{thm}{Theorem}
\newtheorem{prop}{Proposition}

\newtheorem{crl}{Corollary}
\theoremstyle{definition}
\newtheorem{defn}{Definition}
\newtheorem{ex}{Example}

\newcommand{\eg}{\textit{e.g.}}
\newcommand{\ie}{\textit{i.e.}}

\usepackage[skins,breakable]{tcolorbox}

\newtcolorbox{promptbox}[1][]{
  colback=gray!10,%
  colframe=gray!50,%
  coltitle=black,%
  boxrule=0.8pt,
  arc=6pt,
  left=6pt,
  right=6pt,
  top=6pt,
  bottom=6pt,
  fonttitle=\bfseries,
  title=#1,
  fontupper=\ttfamily\raggedright
}

\usepackage{adjustbox}
\usepackage{array}

\newcolumntype{R}[2]{%
    >{\adjustbox{angle=#1,lap=\width-(#2)}\bgroup}%
    l%
    <{\egroup}%
}
\newcommand*\rot{\multicolumn{1}{R{45}{1em}}}%

\title{When Guessing is Rewarded: \\ Rethinking Language Model Evaluation with \\ Distributional Uncertainty Scoring}

\author{%
  Thomas F Burns\\
  Aleph Alpha
}

\begin{document}

\maketitle

\begin{abstract}
Standard language model evaluation assigns scores to single predicted answers, rewarding high-confidence responses regardless of how residual probability mass is distributed over alternative options. This creates a systematic pressure toward overconfident guessing: under accuracy-based schemes, a model maximises its expected score by always committing to an answer rather than abstaining, even when its uncertainty is high. While penalty-based approaches partially address this by raising the confidence threshold for strategic guessing, they still treat all sub-threshold responses identically, ignoring a fundamental distinction in how models can express uncertainty --- for example between hedging toward incorrect answers versus hedging toward ``I don't know'' responses. This paper introduces a novel evaluation metric to solve this problem of not considering a model's entire probability distribution over answer choices. The metric naturally distinguishes between harmful overconfidence in wrong answers and uncertainty expressed through abstention, providing scores in an interpretable default range. Through theoretical analysis and illustrative examples, the metric is shown to offer a more nuanced and aligned evaluation paradigm that incentivises models to express genuine uncertainty rather than guessing. Adapting 12 existing evaluation benchmarks to the metric's variants and measuring performance on six language models shows that for half of the tested benchmarks scores are \textit{negative across all tested models}, indicating significant tendencies towards hallucination.
\end{abstract}

\section{Introduction}\label{sec:intro}

Evaluation of language models has commonly focused on whether they produce `correct' or desired outputs in response to given inputs or instructions, as measured using accuracy or probability-based scoring rules that account for confidence in model predictions. However, the paradigm of focusing on a single answer fundamentally misses a critical aspect of evaluating performance: how models distribute their beliefs across the space of possible responses, including the possibility of abstaining from answering in conditions of uncertainty.

Recent work \citep{kalai2025languagemodelshallucinate} provides compelling evidence that language model `hallucinations' persist in-part due to the socio-technical problem of flawed evaluation metrics. Under traditional binary scoring -- where correct answers receive a positive score (maximally $1$ for perfect correctness), any response like ``I don't know'' (IDK) receives 0, and incorrect answers also receive 0 -- the optimal strategy for any rational agent is to always guess rather than abstain, even when confidence in the guess is minimal. This creates a systematic bias in current evaluation paradigms toward overconfident responses and offers a socio-technical explanation for why language models persist in making confident assertions about uncertain information, \ie, `hallucinate', and possibly engage in unsafe behaviours. This aligns with prior work observing that detrimental phenomena such as hallucination render accuracy an imperfect evaluation metric \citep{hu2024unveiling}. It also aligns with more mechanistic or technical accounts of the causes of hallucination, such as model over-confidence \citep{yin2023largelanguagemodelsknow} and bias towards generating outputs based on common patterns seen in the training corpus (but incorrect in the hallucinated case) while neglecting uncommon patterns (but which could have been correct) \citep{sun2025llmshallucinateconnectingdots}.

In an effort to begin addressing the socio-technical part of this issue, \cite{kalai2025languagemodelshallucinate} propose updating evaluation metrics to penalise incorrect answers, thereby implicitly increasing the optimal confidence threshold required for strategic guessing. However, this kind of threshold-based approach still treats all responses above or below the threshold identically. More critically, it ignores a fundamental distinction in how models can express uncertainty, such as \textit{hedging toward incorrect answers} versus \textit{hedging toward abstention}.

Consider two models answering a multiple-choice question where C is the correct answer and the model's probabilities over the set of possible answers (A, B, C, D, or IDK) are:
\begin{align}\label{ex1}
\text{Model 1:} &\quad \text{A: 31\%, B: 27\%, C: 40\%, D: 1\%, IDK: 1\%} \nonumber \\
\text{Model 2:} &\quad \text{A: 15\%, B: 5\%, C: 40\%, D: 1\%, IDK: 39\%} \nonumber
\end{align}

Both models are equally confident in the correct answer (40\%), yet their belief states represent fundamentally different epistemic positions. Model 1 distributes most of its uncertainty among incorrect options -- a form of `confident incorrectness'. Model 2 hedges primarily toward abstention, expressing considerable uncertainty about the question. Common evaluation schemes which use accuracy as a metric do not distinguish between these qualitatively different types of uncertainty. Given recent causal evidence that language models can use internally-generated confidence signals to drive their own behaviours \citep{kumaran2026causal}, using model-derived uncertainty signals in scoring may be a valid and effective way to achieve fairer, more sensitive, and more roboust evaluation.

This work introduces the \textbf{Distributional Correctness Score (DCS)}, a novel metric that evaluates a model's entire belief distribution rather than just its top prediction(s) or confidence in correctness. The key contributions are:

\begin{enumerate}
    \item characterising the limitations of existing evaluation metrics in capturing model belief states, particularly their failure to distinguish between different types of uncertainty;
    \item introducing DCS, a theoretically grounded metric that, with default settings, produces interpretable scores in $[-1, 1]$ while naturally incorporating the role of abstention as a neutral anchor at $0$;
    \item demonstrating through theoretical analysis that DCS incentivises the desired behaviour: confidence in correct answers, uncertainty when knowledge is lacking, and preference for abstention over confident incorrectness; and
    \item adapting 12 existing benchmarks to use DCS and evaluating six language models, revealing that many language models exhibit systematic epistemic overconfidence, with half the benchmarks showing universally negative DCS scores across all models, the best-performing model achieving only 0.19 DCS (compared to 0.678 accuracy) on its strongest benchmark, and particularly concerning performance gaps on safety-critical benchmarks like TruthfulQA and Winogender.
\end{enumerate}

\section{Related Work}

There is a growing ecosystem of hallucination detection methods and mitigation strategies. Benchmarks such as the Holistic Evaluation of Language Models \citep{liang2022helm} and HaluEval \citep{li2023halueval} provide standardised settings to measure model reliability and truthfulness, fostering reproducible evaluations of hallucination phenomena \citep{ji2022survey,cossio2025taxonomy}. There has also been work on sophisticated interactive-proof-style supervision of information retrieval systems that uses retrieved information as verifiable evidence \citep{deiseroth2026boundinghallucinationsmerlinarthurprotocols}. Within this landscape, detection paradigms have diversified, including sampling-based self-consistency checks such as SelfCheckGPT \citep{manakul2023selfcheckgpt}, semantic-uncertainty estimators \citep{farquhar2024semantic}, model parameter perturbation \citep{liu2025enhancinghallucinationdetectionnoise}, and internal-state probes that read hidden activations \citep{azaria2023internal}. The approach introduced in this work is closer to token-probability approaches \citep{quevedo2024token}, but differs conceptually: it connects a fundamental cause of hallucinations, which might be called `evaluation pressure', to a principled, observable metric for integration with general and pre-existing benchmarks.

An objective of this work is to help highlight and correct for the fact that current benchmark metrics unfairly reward models which hallucinate. One way to discourage such behaviour is to design specific benchmarks (like mentioned above) which seek to detect and punish hallucination. This work instead takes a more systematic approach by adjusting the metric used across existing and future benchmarks, \ie, rather than creating new benchmarks specifically for hallucination detection, the aim is to more accurately score performance on benchmarks \textit{generally}. This has the advantage of measuring hallucination in `naturally-occurring' ecological or pragmatic contexts (\eg, in existing benchmarks), rather than in isolation, which may otherwise harm validity. This appears especially important methodologically, given evidence that some models may be able to detect when and how they are being evaluated \citep{needham2025largelanguagemodelsknow}.

DCS builds directly on \citet{kalai2025languagemodelshallucinate}'s insight that current evaluation schemes incentivise overconfident guessing. While their penalty-based approach addresses the threshold problem, DCS provides a more granular solution that captures the full richness of model belief states while naturally maintaining the neutral value of abstention. Although DCS is offered as a more robust replacement to existing single-answer accuracy metrics, it also complements several existing metrics, \eg, F1 scores \citep{chinchor1993muc} and the Matthews correlation coefficient \citep{matthews1975comparison} excel at measuring classification performance across multiple categories. Different to such metrics, however, DCS specifically targets the evaluation of model performance under uncertainty in question-answer settings. In this sense, it can be interpreted as a specifically-tailored expected cost \citep{ferrer2025no} for language model evaluation in question-answer contexts.

Our approach also connects to the broader literature on proper scoring rules \citep{gneiting2007strictly}, but extends beyond traditional applications by explicitly modelling the abstention option and distinguishing between different patterns of uncertainty expression over the full response space. While DCS shares goals with proper scoring rules like the Brier score \citep{glenn1950verification} in that it evaluates the entire probability distribution, its objective is different. Proper scoring rules are designed to elicit calibrated event probabilities. DCS, in contrast, is designed to incentivise desired behaviour by implementing asymmetric cost functions which include an explicit IDK option.

\section{The Limits of Single-Answer Accuracy Evaluations}\label{sec:limitations-of-accuracy}

Traditional accuracy metrics which use (proxies of) model confidence, \eg, log-likelihood values, operate by taking the \textsc{argmax} of a model's probability output distribution\footnote{Either in whole or in the subset of outputs corresponding to possible answers, \eg, ``A'', ``B'', ``C'', or ``D'' in a four-way multiple-choice question.} and checking if it matches the ground truth answer $c$ to an evaluated question. A variation of this approach is to instead use the confidence itself as a score, \ie, $s = p_{\text{c}}$, where $s$ is the score and $p_{\text{c}}$ is the probability of the model generating the correct answer as its next output. Completion-based implementations of accuracy are even simpler: let the model generate an answer and perform some form of matching, \eg, token or string matching, between its output and the ground truth.

However, all of these approaches have problems. The \textsc{argmax} and completion-based approaches discard information about the model's (relative) uncertainty, whereas using the confidence itself as a score treats vastly different global model belief states identically. To help illustrate the former, consider the following answer probability distributions for a four-way multiple choice question where option C is correct:
\begin{align}
\text{Model 1:} &\quad \text{A: 1\%, B: 1\%, C: 97\%, D: 1\%} \nonumber \\
\text{Model 2:} &\quad \text{A: 24\%, B: 25\%, C: 26\%, D: 25\%} \nonumber
\end{align}

Using the \textsc{argmax} both models would receive identical accuracy scores of $1$ (perfectly correct). Yet, Model 1 demonstrates strong, justified confidence while Model 2 represents a barely-better-than-random guess. This failure to distinguish confidence levels facilitates a disincentive for models to be genuinely certain (or otherwise express uncertainty) versus `gaming' evaluations, \ie, by guessing.

Using the confidence itself as a score is also problematic: while it does not ignore the quantity of remaining probability mass, it ignores how that mass is distributed, so (as in the first example of §\ref{sec:intro}) it cannot distinguish models with identical confidence in the correct answer that spread their uncertainty very differently. The most critical limitation of single-answer accuracy metrics is therefore their inability to capture \textit{that} and \textit{how} models express uncertainty.

\section{The Distributional Correctness Score}\label{sec:dcs}

To address these limitations, the Distributional Correctness Score (DCS) evaluates a model's entire probability distribution over answer choices, including an IDK response option.

Let the model's output assign probabilities to the set of answers $\mathcal{A}$, which includes subsets of correct answers $C$, incorrect answers $W$, and IDK answers $K$. Without loss of generality, questions are assumed henceforth to have a single correct answer, $c$, and a single IDK response, $\textsc{idk}$.

\begin{defn}[Distributional Correctness Score]\label{def:DCS}
The DCS of a model $\mathcal{M}$ is calculated by
\begin{equation*}
    \text{DCS}_{\mathcal{M}} (p_c, P_{W}, p_{\textsc{idk}}, l_c, l_w):= (l_c p_c - l_w P_W) \cdot (1 - p_{\textsc{idk}}),
\end{equation*}
where $p_c$ is the probability assigned to the correct answer, $P_W$ is the sum of probabilities assigned to all incorrect answers, $p_{\textsc{idk}}$ is the probability assigned to ``I don't know'' or a similar abstention response, $l_c \geq 0$ is the loading of the correct response, $l_w \geq 0$ is the loading of the incorrect response(s), and $l_c \geq l_w$.
\end{defn}

The term $(l_c p_c - l_w P_W)$ captures the balance between belief in correctness (weighted by $l_c$) versus incorrectness (weighted by $l_w$). For simplicity and unless stated otherwise, $l_c = l_w = 1$, which gives an interpretable, symmetric range from $-1$ (perfectly incorrect) to $+1$ (perfectly correct)\footnote{The loading terms ($l_c$ and $l_w$) are included to: (i) align with prior work \citep{kalai2025languagemodelshallucinate} which provide implied modifiable confidence thresholds and variable penalties for incorrect answers; and (ii) provide users with the explicit ability to set desired penalties and rewards. This is further discussed in §\ref{sec:theory} and §\ref{sec:discussion}.}. The IDK damping factor $(1 - p_{\textsc{idk}})$ pulls the score toward zero in proportion to the model's expressed uncertainty, reflecting the neutral value of abstention, such that if $p_{\textsc{idk}}=1, DCS=0$.

The following examples show how DCS distinguishes between various model strategies.

\begin{ex}[Error-Hedging vs. Abstention-Hedging.] Consider the motivating example where C is correct.
\begin{itemize}
    \item \textbf{Error-Hedging Model:} A: 31\%, B: 27\%, C: 40\%, D: 1\%, IDK: 1\%
    \begin{equation*}
        \text{DCS} = (0.40 - (0.31 + 0.27 + 0.01)) \cdot (1 - 0.01) = \mathbf{-0.1881}
    \end{equation*}
    
    \item \textbf{Abstention-Hedging Model:} A: 15\%, B: 5\%, C: 40\%, D: 1\%, IDK: 39\%
    \begin{equation*}
        \text{DCS} = (0.40 - (0.15 + 0.05 + 0.01)) \cdot (1 - 0.39) = \mathbf{+0.1159}
    \end{equation*}
\end{itemize}
\end{ex}

The Error-Hedging Model distributes uncertainty among incorrect options, suggesting confusion about which specific answer is correct while maintaining confidence that some answer is (at least partially) known. Whereas, the Abstention-Hedging model direct uncertainty toward IDK responses, expressing what we might describe a `epistemic humility', a desirable behaviour.

These patterns have vastly different implications for trustworthiness and reliability. A model that hedges toward errors when uncertain is more dangerous in safety-critical deployment (and arguably less useful in general applications) than one that hedges toward abstention, yet current metrics treat them identically or even favour error-hedging models that happen to guess correctly.

\begin{ex}[Lucky Guesses vs. Confident Knowledge.] Traditional metrics fail to distinguish lucky guesses from genuine knowledge. Consider the following examples where C is correct.
\begin{itemize}
    \item \textbf{Lucky Model:} A: 25\%, B: 24\%, C: 26\%, D: 24\%, IDK: 1\%
    \begin{equation*}
        \text{DCS} = (0.26 - (0.25 + 0.24 + 0.24)) \cdot (1 - 0.01) = \mathbf{-0.4653}
    \end{equation*}
    
    \item \textbf{Confident Knowledge Model:} A: 1\%, B: 1\%, C: 96\%, D: 1\%, IDK: 1\%
    \begin{equation*}
        \text{DCS} = (0.96 - (0.01 + 0.01 + 0.01) \cdot (1 - 0.01) = \mathbf{+0.9207}
    \end{equation*}
\end{itemize}
\end{ex}

DCS appropriately penalises the lucky guess with a negative score while rewarding knowledgeable confidence with a score near $1$.

\section{Theoretical Analysis}\label{sec:theory}

DCS is designed to produce interpretable scores while creating a natural incentive hierarchy that encourages trustworthy behaviour. These properties are formalised here, with all proofs provided in Appendix \ref{app:proofs}.

\begin{thm}[Score Bounds \& Incentive Ordering]\label{thm:score-bounds-and-incentive-ordering}
DCS, as per Definition \ref{def:DCS}, under the assumption of default loadings ($l_c = l_w = 1$), is bounded in the range $[-1,1]$ for any valid probability distribution. Let $\pi = (p_c, P_W, p_{\textsc{idk}})$ be such a distribution. Consider the following three canonical distributions:
\begin{enumerate}
    \item $\pi_{CC} = (1, 0, 0)$ (Confident Correctness).
    \item $\pi_{HA} = (0, 0, 1)$ (Honest Abstention).
    \item $\pi_{CI} = (0, 1, 0)$ (Confident Incorrectness).
\end{enumerate}
Then, $\text{DCS}(\pi_{CI}) < \text{DCS}(\pi_{HA}) < \text{DCS}(\pi_{CC})$.
\end{thm}

The bounded and symmetric range (when using the default loadings) given by Theorem \ref{thm:score-bounds-and-incentive-ordering} makes DCS scores easy to interpret. More importantly, the score structure incentivises a clear preference ordering over epistemic states. In this way, DCS can also be understood as implementing a specific expected cost structure \citep{ferrer2025no} that penalises confident wrongness more severely than uncertain neutrality. This cost structure naturally emerges from many real-world scenarios where overconfident errors cause more harm than appropriate uncertainty, such as safety-relevant applications, \eg, in healthcare \citep{burns2023epistemic,burns2024position,Bedi2025}.

The neutral value of abstention in DCS also aligns with desirable information-theoretic properties. For queries where the training data provides insufficient information to form a confident belief, any model that generates a high-confidence, non-abstaining response is necessarily hallucinating. DCS correctly penalises this behaviour. By rewarding models that map low epistemic certainty to distributions with a score near zero (through high $p_{\textsc{idk}}$ and/or $P_W$), DCS incentivises a more epistemically honest reflection of the model's underlying knowledge. This leads to a preference for models that, if they are to  `hedge their bets', they should do so only to the degree they are truly confident of a particular piece of knowledge, and otherwise abstain from answering, as shown in Corollary \ref{crl:absence-hedging-pref}.

\begin{crl}[Preference for Abstention-Hedging]\label{crl:absence-hedging-pref}
Let $\pi_1 = (p_c, P_{W1}, p_{\textsc{idk1}})$ and $\pi_2 = (p_c, P_{W2}, p_{\textsc{idk2}})$ be two probability distributions over the answer space such that they satisfy the following conditions:
\begin{enumerate}
    \item They assign the same probability to the correct answer, with $0<p_c<1$.
    \item Distribution $\pi_1$ is more confident in incorrectness than abstention $P_{W1} > p_{\textsc{idk1}}$, whereas $\pi_2$ is more confident in abstention than incorrectness $p_{\textsc{idk2}} > P_{W2}$.
    \item The total probability assigned to the answer space $\mathcal{A}$ is equal.
\end{enumerate}
Then, the DCS of $\pi_2$ (abstention-hedging) is strictly greater than of $\pi_1$ (error-hedging).
\end{crl}

Although DCS is defined directly on the model's unconditional output probabilities, it can be rewritten as a two-stage mixture over an implicit abstention decision. First, notice that for an answer $a$, its total probability, $p_a$, can be decomposed into two parts
\begin{equation}\label{eq:prob-substitution}
    p_a= p_{\textsc{idk}} \overline{b_a} + (1-p_{\textsc{idk}}) b_a,
\end{equation}
where $\overline{b_a}$ is the probability of answering $a$ given the model \textit{doesn't believe} it knows the answer, and $b_a$ is the probability of the model answering $a$ given the model \textit{does believe} it knows the answer. If a model has a state where $\overline{b_a} \approx 1$ for the IDK response and $\overline{b_a} \approx 0$ for all other responses in the answer set, the model is said to have a \textit{pure IDK state}.

Definition \ref{def:DCS} can then be reformulated using the substitution from Equation \ref{eq:prob-substitution} to give:
\begin{equation*}
    \text{DCS} = \left[ l_c ( p_{\textsc{idk}} \overline{b_c} + (1-p_{\textsc{idk}}) b_c ) - l_w ( p_{\textsc{idk}} \overline{B_W} + (1-p_{\textsc{idk}}) B_W ) \right] \cdot (1 - p_{\textsc{idk}}),
\end{equation*}
where $\overline{b_c},b_c$ are the conditional probabilities of answering with the correct answer, $c$, given the model doesn't or does (respectively) believe it knows the answer, and where $\overline{B_W},b_w$ are the sum of conditional probabilities for the same but for the incorrect answers, $W$. In a pure IDK state, DCS simplifies to $\sim [l_c b_c - l_w B_W] (1 - p_{\textsc{idk}})$. This lets DCS be described as evaluating the quality of the answering policy inside the `not-IDK' branch, weighted by the probability of entering that branch.

\begin{prop}[Optimal Guessing Threshold]\label{prop:optimal-guessing-threshold}
Suppose a rational agent has a probability $p_c^* \in (0, 1]$ that its most likely answer is correct. The agent is only rewarded for providing a correct answer via an output distribution $\pi = (p_c^*, 1-p_c^*, 0)$ if its score is greater than the abstention score of 0. This is true if and only if its confidence $p_c^*$ exceeds a specific threshold determined by the loadings:
$$p_c^* > \frac{l_w}{l_c + l_w}$$
Under the default symmetric loadings ($l_c = l_w = 1$), this threshold is $p_c^* > 0.5$.
\end{prop}

\begin{wrapfigure}{r}{0.5\textwidth}
  \vspace{-.7cm}
  \begin{center}
    \includegraphics[width=0.42\textwidth]{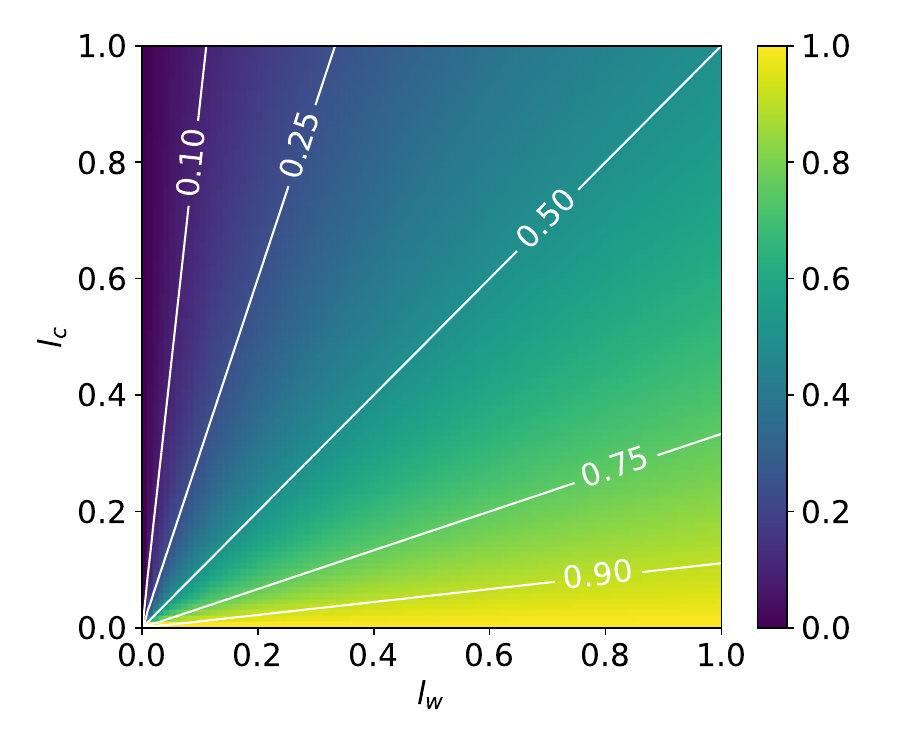}
  \end{center}
  \vspace{-0.6cm}
  \caption{Values of the optimal guessing threshold in DCS, as given by Proposition \ref{prop:optimal-guessing-threshold} for different values of the parameters $l_c$ and $l_w$.}\label{fig:optimal-guessing-thresholds}
  \vspace{-0.6cm}
\end{wrapfigure}

Proposition \ref{prop:optimal-guessing-threshold} provides more interpretable control over DCS, since the optimal guessing threshold is now known for given values of the loadings $l_c$ and $l_w$. As shown in Figure \ref{fig:optimal-guessing-thresholds}, any threshold in the range $(0,1)$ may be chosen.

Given a desired guessing threshold, $p_c^* \in (0,1),$ choose loadings so that $l_w/l_c=p_c^*/(1-p_c^*)$. A convenient default is to set $l_c=1, l_w=p_c^*/(1-p_c^*)$. Then, to achieve, for example, a desired threshold of $0.1$, set $l_w=1/9$; for a threshold of $0.75$, set $l_w=3$; and so on. For convenience, a table of values for setting $l_w$ given $l_c=1$ for some desired guessing thresholds $p_c^*$ is provided in Appendix \ref{app:desired-thresholds}.

\begin{prop}[Information-Theoretic Performance Bound]\label{prop:information-theoretic-bound}
Let $\mathfrak{Q}$ be a random variable representing a query, $\mathfrak{A}$ be the random variable for its correct answer over a set of size $k$, and $\mathcal{D}$ be the training data. The maximum expected DCS achievable by any model $\mathcal{M}$ is upper-bounded by a function of the mutual information between the answer and the data, conditioned on the query:
$$\max_{\mathcal{M}} \mathbb{E}_{\mathfrak{Q};\mathcal{D}}[DCS_\mathcal{M}(\pi(\mathfrak{Q};\mathcal{D})] \le f(I(\mathfrak{A}; \mathcal{D} | \mathfrak{Q})),$$
where $\pi(\mathfrak{A}|\mathfrak{Q};\mathcal{D})$ represents the probability vector that model $\mathcal{M}$ produces for query $\mathfrak{Q}$ given training data $\mathcal{D}$, and where $f$ is a monotonically increasing function such that as the conditional mutual information $I(\mathfrak{A}; \mathcal{D} | \mathfrak{Q}) \to 0$, the maximum expected score $f(I) \le 0$ for $k>2$ and $f(I) = 0$ for $k=2$.
\end{prop}

Proposition \ref{prop:information-theoretic-bound} implies that even if a model perfectly optimises its output probabilities to maximise DCS, its expected score is ultimately limited by the conditional mutual information $I(\mathfrak{A};\mathcal{D}|\mathfrak{Q})$: if the training data carry little or no information about the answer beyond the query itself, the best achievable DCS collapses toward zero. The monotonicity of $f$ formalises the intuition that richer, more informative data can support higher correctness while still accounting for abstention and misallocation of probability mass. Practically, this bound warns practitioners that improvements in architecture or inference cannot substitute for information content in the data  -- when $I(\mathfrak{A};\mathcal{D}|\mathfrak{Q})$ is small, even very expressive models will yield near-chance DCS (and negative values when $k>2$ and incorrect responses dominate). It also provides a principled way to compare tasks: those with larger answer spaces or weaker data-answer dependence inherently admit lower DCS ceilings.

\subsection{Grounding and interpretation}\label{subsec:grounding}

DCS is a product of two separately motivated factors. The linear term $(l_c p_c - l_w P_W)$ is the expected value, under the model's own answer distribution, of the penalty-based score of \citet{kalai2025languagemodelshallucinate}, wherein a correct answer earns $+l_c$, a wrong answer costs $-l_w$, and abstention earns $0$; in the sense of \citet{ferrer2025no}, this is an expected cost with an explicit, tunable cost structure. The second factor $(1 - p_{\textsc{idk}})$ is the weight of the `not-IDK' branch in the two-stage mixture reformulation of Equation \ref{eq:prob-substitution}, wherein DCS evaluates the answering policy inside that branch, weighted by the probability of entering it. The loadings ratio, the only remaining free choice, is fixed by Proposition \ref{prop:optimal-guessing-threshold} through a target confidence threshold $p_c^* = l_w/(l_c + l_w)$.

DCS is not claimed to be axiomatically unique, and it is intentionally not a proper scoring rule. Unlike forecasting with stochastic labels, language model evaluations present deterministic facts of the matter, so the `true' conditional distribution is a point mass and it is meaningless to demand a model report the frequency of correctness per option (as the Brier score rewards). Proper scoring rules elicit calibrated credences, whereas DCS targets a decision-theoretic objective: an asymmetric utility that prices confident wrongness below abstention. The two are complementary: the term $(l_c p_c - l_w P_W)$ explicitly penalises `confident incorrectness' (high $P_W$) more than `uncertainty' (high $p_{\textsc{idk}}$), a distinction crucial for mitigating harmful hallucinations but not the focus of traditional calibration.

DCS treats several belief states as equivalent. The level sets are $\{(p_c, P_W, p_{\textsc{idk}}) : (l_c p_c - l_w P_W)(1 - p_{\textsc{idk}}) = v\}$, so the score-zero surface is two-dimensional rather than a single state, containing honest abstention ($p_{\textsc{idk}} = 1$), balanced confusion ($l_c p_c = l_w P_W$), and any mixture of the two. Collapsing these to the same neutral value is intended, since each represents an absence of net correct signal. Within a fixed $p_{\textsc{idk}}$, the score depends on $(p_c, P_W)$ only through $l_c p_c - l_w P_W$; DCS is thus deliberately blind to how a fixed amount of wrong-answer mass is spread across distractors, a modelling choice one could readily generalise to weight different incorrect answers differently where a deployment context calls for it.

\section{Experiments}\label{sec:experiments}

Six language models were evaluated. These models were: DialoGPT-Medium, Llama3.2 3B Instruct, Llama TFree HAT Pretrained 7B DPO, Mistral 7B Instruct v0.3, Llama3.1 8B Instruct, and DeepSeek R1 0528 Qwen3 8B. Brief overviews of each of these models are provided in Appendix \ref{app:evaluated-models}. All inference was performed using \textsc{transformers}\footnote{\url{https://github.com/huggingface/transformers}} and evaluated using \textsc{eval-framework}\footnote{\url{https://github.com/Aleph-Alpha-Research/eval-framework}}. Appendix~\ref{app:licenses} summarises licence identifiers for software, checkpoints, and benchmark corpora.

The DCS metric was implemented for 12 established benchmarks, namely: ARC, COPA, GPQA, HellaSwag, MMLU, MMLU Pro, PIQA, OpenBookQA, SciQ, TruthfulQA, Winogender, and Winogrande. For completeness, a brief description of each of these benchmarks is provided in Appendix \ref{app:evaluated-benchmarks}. For each benchmark, examples and the prompting template are provided in Appendix \ref{app:prompting-templates}. In each case, the log-likelihoods of the standard answer set $\mathcal{A}$ are computed, as well as an IDK response.

The quantities $(p_c, P_W, p_{\textsc{idk}})$ are obtained as follows. Each candidate completion in the option set (the standard answers plus a length-matched IDK completion) is scored by its length-normalised log-likelihood, i.e., the sum of its token log-probabilities divided by its token length, conditioned on the shared prompt. A softmax over these values then yields a distribution from which $p_c$, $P_W$, and $p_{\textsc{idk}}$ are read directly, with $p_{\textsc{idk}}$ the mass on the IDK completion. Length normalisation avoids a systematic length bias when comparing options of differing lengths, including the added IDK option, constructed to roughly match the others in length and formatting (Appendix \ref{app:prompting-templates}).

To better understand and compare the performance of DCS, the accuracy, confidence-weighted accuracy, and the proposed metric of \cite{kalai2025languagemodelshallucinate} (referred to as the ternary score) were also computed for each model inference completion. Table \ref{tab:metrics} (Appendix \ref{app:other-metrics}) gives the exact definition of each comparison score, $s$. To aid readability, all scores are multiplied by $100$.

\begin{figure}[h]
    \centering
    \includegraphics[width=0.52\linewidth]{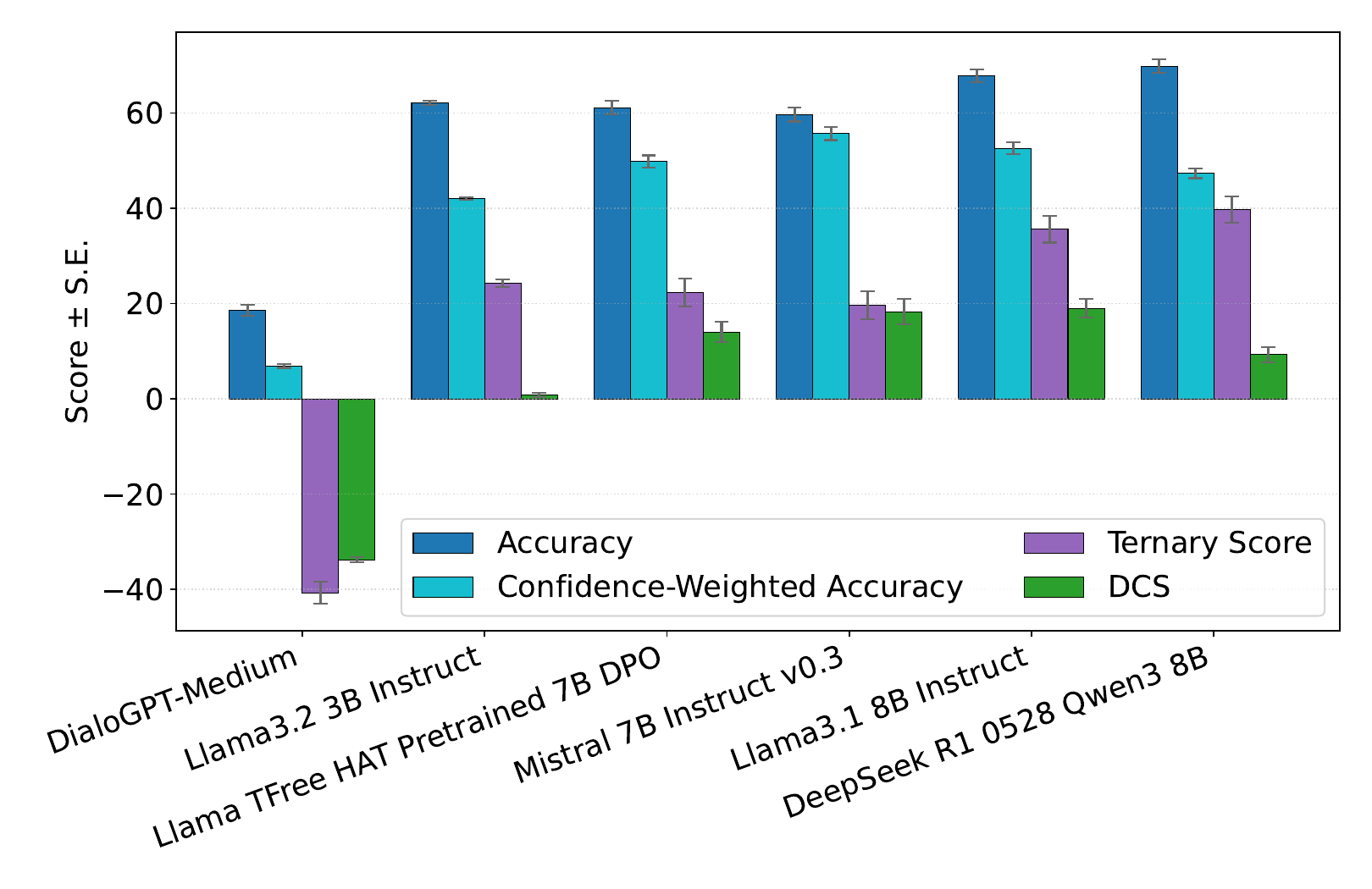}
    \caption{Mean MMLU scores as measured by accuracy, confidence-weighted accuracy, ternary score, and DCS across the evaluated models. All scores are multiplied by $100$ for readability. Error bars represent the standard error (S.E.).}
    \label{fig:mmlu}
\end{figure}

Figure \ref{fig:mmlu} shows results for MMLU, comparing the six models on each of the metrics in Table \ref{tab:metrics} and the DCS. In all cases, the DCS is significantly lower than both the accuracy and confidence-weighted accuracy metrics. The DCS is also significantly different to the ternary score across all models (unpaired t-test, $p<0.0001$), including Mistral 7B Instruct v0.3, where the DCS appears very similar but is actually lower (unpaired t-test, $p<0.0001$, $t=10.5548$; the $95\%$ confidence interval of this difference is from $-1.542$ to $-1.058$).

However, mean DCS values are not always lower than the ternary score, as the comparison for DialoGPT-Medium in Figure \ref{fig:mmlu} shows. Here, the ternary score is significantly lower than the DCS. This reflects the fact that DCS does not only punish models which do not assign a larger probability mass to the correct answer, but also rewards models which -- even when wrong -- hedge towards IDK or correct responses. Results from other benchmarks show similar trends.

Table \ref{tab:DCS-all-results} shows mean DCS values ($\pm$ S.E.) across all models and benchmarks (see Appendix \ref{app:other-metrics} for the other metrics). Three findings stand out: (1) for half of the tested benchmarks, DCS scores are \textit{negative across all tested models}; (2) the highest mean DCS for any benchmark and model was 0.19 (Llama3.1 8B Instruct on MMLU), less than a third of its traditionally-measured accuracy of 0.678; and (3) benchmarks important for trust and safety, such as TruthfulQA and Winogender, are among those with negative DCS or a large accuracy--DCS gap. 

\begin{table}[h]
    \centering
    \small\setlength{\tabcolsep}{4pt}\renewcommand{\arraystretch}{0.9}
    \begin{tabular}{lcccccc}
                      & \rot{DialoGPT-Medium} & \rot{Llama3.2 3B Instruct} & \rot{Llama TFree HAT Pretrained 7B DPO} & \rot{Mistral 7B Instruct v0.3} & \rot{Llama3.1 8B Instruct} & \rot{DeepSeek R1 0528 Qwen3 8B} \\
        ARC           & -24.5 $\pm$ 0.3 & -15.1 $\pm$ 0.2 & \underline{-12.1 $\pm$ 0.3} & \textbf{-10.6 $\pm$ 0.3} & -13.9 $\pm$ 0.3 & -19.6 $\pm$ 0.3 \\
        COPA          & 0.6 $\pm$ 0.4 & 2.9 $\pm$ 0.4 & \underline{3.4 $\pm$ 0.4} & \textbf{10.2 $\pm$ 1.0} & 3.0 $\pm$ 0.3 & 3.2 $\pm$ 1.1 \\
        GPQA          & \textbf{-21.6 $\pm$ 0.4} & -44.6 $\pm$ 0.7 & -43.0 $\pm$ 1.1 & -44.7 $\pm$ 1.6 & -41.9 $\pm$ 1.0 & \underline{-39.3 $\pm$ 0.5} \\
        HellaSwag     & -34.0 $\pm$ 0.2 & -26.2 $\pm$ 0.1 & \underline{-23.2 $\pm$ 0.2} & \textbf{-19.0 $\pm$ 0.2} & -27.8 $\pm$ 0.2 & -37.2 $\pm$ 0.3 \\
        MMLU          & -33.8 $\pm$ 0.5 & 0.8 $\pm$ 0.5 & 14.0 $\pm$ 2.1 & \underline{18.3 $\pm$ 2.6} & \textbf{19.0 $\pm$ 1.9} & 9.3 $\pm$ 1.6 \\
        MMLU Pro      & -61.5 $\pm$ 0.3 & -56.8 $\pm$ 0.3 & \underline{-38.8 $\pm$ 1.9} & \textbf{-34.6 $\pm$ 2.6} & -38.9 $\pm$ 1.8 & -40.3 $\pm$ 1.4 \\
        OpenBookQA    & -20.3 $\pm$ 0.5 & -15.9 $\pm$ 0.5 & \textbf{-11.5 $\pm$ 0.5} & \underline{-12.7 $\pm$ 0.5} & -14.9 $\pm$ 0.6 & -21.3 $\pm$ 0.8 \\
        PIQA          & 0.2 $\pm$ 0.1 & 1.4 $\pm$ 0.1 & \underline{1.9 $\pm$ 0.1} & \textbf{3.3 $\pm$ 0.2} & 1.4 $\pm$ 0.1 & 1.4 $\pm$ 0.1 \\
        SciQ          & -21.0 $\pm$ 0.4 & 1.4 $\pm$ 0.5 & 4.0 $\pm$ 0.4 & \textbf{9.0 $\pm$ 0.6} & \underline{7.9 $\pm$ 0.5} & -6.3 $\pm$ 0.4 \\
        TruthfulQA    & -43.6 $\pm$ 0.4 & -40.1 $\pm$ 0.4 & \underline{-36.6 $\pm$ 0.4} & \textbf{-33.8 $\pm$ 0.4} & -39.4 $\pm$ 0.4 & -43.9 $\pm$ 0.4 \\
        Winogender    & 0.1 $\pm$ 0.6 & 1.4 $\pm$ 0.3 & 2.2 $\pm$ 0.6 & \textbf{18.5 $\pm$ 1.2} & 1.4 $\pm$ 0.2 & \underline{2.5 $\pm$ 1.3} \\
        Winogrande    & -0.1 $\pm$ 0.1 & 0.8 $\pm$ 0.1 & 1.1 $\pm$ 0.1 & \underline{1.3 $\pm$ 0.1} & \textbf{1.4 $\pm$ 0.1} & 0.7 $\pm$ 0.2 \\
        \hline
        Average       & -21.63 & -15.83 & -11.55 & \textbf{-7.90} & -11.89 & -15.90  \\
    \end{tabular}
    \caption{Mean DCS ($\pm$ S.E.) across tested benchmarks and models. All scores are multiplied by $100$ for readability. The best (highest) DCS per benchmark is shown in \textbf{bold} and the second best is \underline{underlined}.}\label{tab:DCS-all-results}
\end{table}

Table \ref{tab:deltas} reports, for each benchmark, the mean gap between DCS and each comparison metric averaged across the six models, so the reader need not cross-reference Table \ref{tab:DCS-all-results} against the appendix tables (Appendix \ref{app:other-metrics}). DCS is uniformly lower than accuracy, confidence-weighted accuracy, and the ternary score, often substantially so; the largest gaps fall on knowledge-heavy benchmarks such as MMLU, MMLU Pro, SciQ, and HellaSwag, where models answer confidently but not always correctly.

\begin{table}[h]
    \centering
    \small\renewcommand{\arraystretch}{0.9}
    \begin{tabular}{lccc}
        Benchmark & $\Delta$Accuracy & $\Delta$C-weighted acc. & $\Delta$Ternary \\
        \hline
        ARC          & -28.9 & -19.7 & -23.8 \\
        COPA         & -14.8 & -4.6  & -8.6  \\
        GPQA         & -66.5 & -49.6 & -7.4  \\
        HellaSwag    & -78.8 & -41.4 & -49.6 \\
        MMLU         & -51.9 & -37.8 & -12.2 \\
        MMLU Pro     & -77.6 & -63.9 & -14.6 \\
        OpenBookQA   & -27.2 & -19.7 & -15.1 \\
        PIQA         & -7.2  & -2.0  & -5.9  \\
        SciQ         & -74.4 & -32.7 & -67.8 \\
        TruthfulQA   & -58.4 & -43.7 & -24.4 \\
        Winogender   & -34.5 & -16.4 & -6.9  \\
        Winogrande   & -24.7 & -9.3  & -6.1  \\
        \hline
        Average      & -45.4 & -28.4 & -20.2 \\
    \end{tabular}
    \caption{Per-benchmark mean gap between DCS and each comparison metric (DCS minus the metric), averaged across the six models. As elsewhere, quantities are multiplied by 100 fo readability. Negative values indicate DCS scores below the comparison metric.}\label{tab:deltas}
\end{table}

Beyond the aggregate gap, DCS yields several findings that the other metrics do not expose:
\begin{itemize}
    \item \textbf{Recovery of signal from 0\% accuracy.} Despite format-matched IDK options and length-normalised log-likelihoods, Llama3.2 3B Instruct and Llama3.1 8B Instruct scored exactly $0\%$ accuracy on COPA and PIQA respectively (Table \ref{tab:accuracy}), indicating that their \textsc{argmax} choices violated the task format. Yet both obtained positive DCS on the same benchmarks (COPA: $+2.9$, $+3.0$; PIQA: $+1.4$, $+1.4$). DCS thus recovers signal from the broader distribution while still flagging the instruction-following deficit, neither of which is visible to accuracy alone.
    \item \textbf{Separation of hedging profiles.} On TruthfulQA all models score negatively, yet DCS separates models that hedge toward wrong answers from those that hedge toward abstention, a distinction the ternary score collapses. Mistral 7B Instruct v0.3 (DCS $-33.8$, ternary $0.1$) differs far more between the two metrics than DeepSeek R1 0528 Qwen3 8B (DCS $-43.9$, ternary $-30.6$); the former appears to confidently distribute mass across distractors rather than abstaining.
    \item \textbf{Sensitivity to stated stakes.} Accuracy is loadings-invariant by construction, so it cannot register whether a model responds to the stated stakes of a task; DCS can, being distribution- and loadings-aware. This is revisited in Section \ref{sec:dcs-behaviour}, where a stated-stakes sweep shows models register the incentive sub-behaviourally even when it barely shifts their abstention.
\end{itemize}

Table \ref{tab:DCS-all-results} uses a version of DCS that assumes a single, canonical IDK response, and thus may under-account for $p_{IDK}$ probability mass. To probe this sensitivity, Table \ref{tab:DCS-multi-idk-results} in Appendix \ref{app:multi-phrase-DCS} recomputes DCS for all six models and benchmarks using an expanded abstention set of 10 phrasings with $p_{IDK} = \sum_{i=1}^{10} p_{IDK_i}$. Scores consistently shift toward zero, as expected when models distribute broader abstention mass, but the core ranking relationships and negative-score patterns remain, indicating that under our initial phrasings, DCS could capture genuine epistemic uncertainty rather than artifacts of those specific phrases. Summed abstention probabilities are still recommended when computational feasible, although a single canonical phrase may suffice in some cases (like this one) for comparative evaluation.

DCS principles also extend beyond discrete multiple choice. For continuous answer spaces (\eg, regression, bounding-box estimation, or event timings), DCS integrates the predicted density over a `correct region' against the density over the incorrect region, with a distinct named IDK region as the neutral anchor. For open-ended generation, a sampling-based instantiation draws $N$ generations per question ($N=16$, temperature $0.7$), canonicalises each into a correct, wrong, or abstention bucket (judging correctness by exact match or an LLM-as-judge check), estimates $(p_c, P_W, p_{\textsc{idk}})$ as sample proportions, and applies the same $(p_c - P_W)(1 - p_{\textsc{idk}})$. On 500 TriviaQA items this free-form DCS tracks expected model capability or quality ($+0.023$ for \texttt{Qwen2.5-1.5B-Instruct} versus $+0.639$ for the stronger \texttt{Llama-3.1-8B-Instruct}, non-overlapping 95\% CIs).

\section{Optimising for and predicting explicit hallucination benchmarks with DCS}\label{sec:dcs-behaviour}

Is DCS also a useful training target and a predictor of more sophisticated hallucination behaviour?

\paragraph{Direct optimisation and abstention.} A \texttt{Qwen2.5-1.5B-Instruct} model was fine-tuned with LoRA ($r{=}16$) over a 20{,}000-question MMLU auxiliary-train corpus (deduplicated against the evaluation set), comparing cross entropy against the damped DCS objective $-(p_c - P_W)(1 - p_{\textsc{idk}})$. Optimising DCS directly improved mean held-out DCS over cross entropy on all four evaluated benchmarks tested (MMLU $+0.115$, ARC $+0.044$, SciQ $+0.022$, TruthfulQA $+0.022$; paired item bootstrap), so the metric appears optimisable in practice. What the model is optimised to predict also appears to shape abstention most directly: IDK relabelling in the style of \citet{kang2025unfamiliar} (base-incorrect items relabelled to the abstention option) raises held-out MMLU abstention from near-zero to $0.397$, implying DCS-aligned abstention is learnable and controllable in some settings.

\paragraph{Predicting hedging and hallucination.} Evaluating on the DCS version of MMLU with the scores explicitly stated in the prompt moved mean $p_{\textsc{idk}}$ upward as the stated wrong-answer penalty increased (the direction implied by Proposition \ref{prop:optimal-guessing-threshold}) for all models. Also, using the DCS version of TriviaQA and testing four multiple-choice scores (DCS, accuracy, ternary, and mean $p_{\textsc{idk}}$), scored by the sampling instantiation of Section~\ref{sec:discussion}, DCS was the strongest predictor (Spearman $\rho = -0.53$) of two explicit hallucination benchmark scores: SQuAD 2.0 unanswerable \citep{rajpurkar-etal-2018-know} and HaluEval.

\section{Limitations}\label{sec:limitations}

This study has several limitations that bound the scope of its claims. First, the main experiments evaluate DCS in a \emph{multiple-choice} setting with an \emph{explicit} abstention option; while continuous-answer and sampling-based free-form instantiations of DCS are also sketched, a thorough validation in open-ended generation is left to future work. Second, the main results tables use a single canonical IDK option phrase; abstention mass can be phrasing-sensitive, which a ten-phrase set in Appendix~\ref{app:multi-phrase-DCS} probed but does not exhaustively characterise. Finally, the evaluation covers six models of up to 8B parameters chosen to run on consumer-grade hardware; larger models may behave differently.

\section{Discussion}\label{sec:discussion}

DCS offers a shift in how language models are evaluated, moving from scoring single responses to evaluating entire belief states. By explicitly incorporating the role of abstention and distinguishing between overconfidence and states of uncertainty, DCS provides a nuanced, hallucination-sensitive evaluation methodology focused on eliciting genuine model knowledge and capabilities.

The loadings $l_c$ and $l_w$ let DCS serve as a family of utility functions under varying risk profiles. Rather than a single DCS score, future work might evaluate models across a spectrum of $(l_c,l_w)$ settings \citep{kalai2025languagemodelshallucinate}, yielding a `behavioural calibration' benchmark: a model scoring well across diverse cost structures, from $l_w \gg l_c$ (high stakes) to $l_c > l_w$ (low stakes), is more plausibly responsive to human-expressed preferences than one optimised for a single fixed notion of correctness.

I hope that DCS helps support a practical path toward evaluation that aligns incentives with honesty and epistemic humility, shifting progress measurement from score-gaming toward trustworthy demonstrations of knowledge.

\subsubsection*{Code and reproducibility}
The DCS metric and the evaluation tasks used for the main experiments are implemented in \textsc{eval-framework} (\url{https://github.com/Aleph-Alpha-Research/eval-framework}, Apache 2.0).

\subsubsection*{Acknowledgements}
I thank my colleagues, especially Koen Oostermeijer, for helpful discussions and feedback. The compute and research time spent by the author was funded by Aleph Alpha Research GmbH.

{\small
\bibliographystyle{iclr2026_conference}
\bibliography{iclr2026_conference}
}

\appendix
\section{Appendix}

\subsection{Licenses and terms of use}\label{app:licenses}
All publicly released software, model checkpoints, and benchmark corpora were used only under the terms stated by each provider (model cards, dataset cards, and repository \texttt{LICENSE} files). Table~\ref{tab:licenses-assets} lists the primary license name or family relied upon; some Hugging Face assets require additional click-through or gated-access conditions, which need to be accepted when downloading weights or data.

\begin{table}[t]
    \centering
    \caption{Primary licenses for major third-party assets used in this work (see also Appendix~\ref{app:evaluated-models} for model-card URLs). Where a benchmark is distributed under several compatible terms depending on mirror, the license cited is that stated on the Hugging Face dataset card or canonical repository at access time.}
    \label{tab:licenses-assets}
    \small
    \begin{tabular}{>{\raggedright\arraybackslash}p{0.34\linewidth}p{0.58\linewidth}}
        \toprule
        Asset & License (summary) \\
        \midrule
        \textsc{transformers} library & Apache License 2.0 (repository \texttt{LICENSE}) \\
        \textsc{eval-framework} & Apache License 2.0 (repository \texttt{LICENSE}) \\
        \midrule
        \multicolumn{2}{l}{\textit{Model checkpoints (Hugging Face model cards)}} \\
        \url{microsoft/DialoGPT-medium} & MIT \\
        \url{meta-llama/Llama-3.2-3B-Instruct} & Llama~3.2 Community License Agreement \\
        \url{meta-llama/Llama-3.1-8B-Instruct} & Llama~3.1 Community License Agreement \\
        \url{mistralai/Mistral-7B-Instruct-v0.3} & Apache License 2.0 \\
        \url{Aleph-Alpha/llama-tfree-hat-pretrained-7b-dpo} & Open Aleph License (non-commercial research and educational use; see model-card link) \\
        \url{deepseek-ai/DeepSeek-R1-0528-Qwen3-8B} & MIT License (model-card \texttt{LICENSE}; Qwen3 upstream under its own terms) \\
        \midrule
        \multicolumn{2}{l}{\textit{Benchmark corpora (typical distribution on Hugging Face / original releases)}} \\
        ARC (AI2) & CC BY-SA 4.0 \\
        COPA & Free for research use (original distribution terms) \\
        GPQA & CC BY~4.0 on Hugging Face; dataset-specific access conditions apply \\
        HellaSwag & MIT License \\
        MMLU & MIT License \\
        MMLU-Pro & MIT License (Hugging Face dataset card, e.g.\ \texttt{TIGER-Lab/MMLU-Pro}) \\
        OpenBookQA, SciQ, PIQA & Apache License 2.0 \\
        TruthfulQA & Apache License 2.0 \\
        Winogender & CC BY 4.0 \\
        Winogrande (WinoGrande) & CC BY 4.0 (SuperGLUE-style release) \\
        \bottomrule
    \end{tabular}
\end{table}

\subsection{Proofs}\label{app:proofs}

\setcounter{thm}{0}
\begin{thm}[Score Bounds \& Incentive Ordering]
DCS, as per Definition \ref{def:DCS}, under the assumption of default loadings ($l_c = l_w = 1$), is bounded in the range $[-1,1]$ for any valid probability distribution. Let $\pi = (p_c, P_W, p_{\textsc{idk}})$ be such a distribution. Consider the following three canonical distributions:
\begin{enumerate}
    \item $\pi_{CC} = (1, 0, 0)$ (Confident Correctness).
    \item $\pi_{HA} = (0, 0, 1)$ (Honest Abstention).
    \item $\pi_{CI} = (0, 1, 0)$ (Confident Incorrectness).
\end{enumerate}
Then, $\text{DCS}(\pi_{CI}) < \text{DCS}(\pi_{HA}) < \text{DCS}(\pi_{CC})$.
\end{thm}
\begin{proof}
Let $f(p_c, P_W, p_{\textsc{idk}}) = (p_c - P_W)(1 - p_{\textsc{idk}})$. The goal is to find the global minimum and maximum of this function subject to a set of constraints.

The variables $p_c, P_W, p_{\textsc{idk}}$ are probabilities or sums of probabilities, so they must be non-negative. Furthermore, the total probability assigned to the answer set $\mathcal{A}$ cannot exceed 1. The domain $\mathcal{D}$ defined by these constraints is a compact set in $\mathbb{R}^3$ (a tetrahedron). The function $f$ is continuous on $\mathcal{D}$. By the Extreme Value Theorem, $f$ must attain a global maximum and minimum on $\mathcal{D}$. These extrema can occur in the interior of $\mathcal{D}$ or on its boundary.

The two factors of $f$ are analysed separately. The term $(p_c - P_W)$ is bounded. Since $p_c \ge 0$ and $P_W \ge 0$, and $p_c + P_W \le 1$, the maximum value of $p_c - P_W$ is $1$ (when $p_c=1, P_W=0$) and the minimum value is $-1$ (when $p_c=0, P_W=1$). So, $-1 \le (p_c - P_W) \le 1$. The term $(1 - p_{\textsc{idk}})$ is bounded. Since $0 \le p_{\textsc{idk}} \le 1$, $0 \le (1 - p_{\textsc{idk}}) \le 1$. The product is therefore also bounded:
$$ -1 \cdot 1 \le (p_c - P_W)(1 - p_{\textsc{idk}}) \le 1 \cdot 1 $$
$$ -1 \le \text{DCS} \le 1 $$
This shows that the score is bounded within $[-1, 1]$.

These bounds are now shown to be sharp, \ie, they can be achieved at specific points in the domain $\mathcal{D}$, and three of these correspond to the stated canonical distributions. Checking the vertices of the tetrahedron $\mathcal{D}$ gives:
\begin{itemize}
    \item Vertex 1: $(p_c, P_W, p_{\textsc{idk}}) = f(0, 0, 0) = (0 - 0)(1 - 0) = 0 $.
    This corresponds to having confidence only in the complement of the answer set $\mathcal{A}'$.
    \item Vertex 2: $\pi_{\textsc{CC}} = f(1,0,0) = (1 - 0)(1 - 0) = 1 $.
    This corresponds to perfect confidence in the correct answer and attains the maximum possible score.
    \item Vertex 3: $\pi_{\textsc{CI}} = f(0,1,0) = (0 - 1)(1 - 0) = -1 $.
    This corresponds to (summed) perfect confidence in the incorrect answer set and attains the minimum possible score.
    \item Vertex 4: $\pi_{\textsc{HA}} = f(0,0,1) = (0 - 0)(1 - 1) = 0 $.
    This corresponds to perfect confidence in abstention.
\end{itemize}
Since the function attains the values of $1$ and $-1$ within the domain $\mathcal{D}$, the bounds are sharp. Direct calculation also shows that $\text{DCS}(\pi_{CI}) < \text{DCS}(\pi_{HA}) < \text{DCS}(\pi_{CC})$.
\end{proof}

\setcounter{crl}{0}
\begin{crl}[Preference for Abstention-Hedging]
Let $\pi_1 = (p_c, P_{W1}, p_{\textsc{idk1}})$ and $\pi_2 = (p_c, P_{W2}, p_{\textsc{idk2}})$ be two probability distributions over the answer space such that they satisfy the following conditions:
\begin{enumerate}
    \item They assign the same probability to the correct answer, with $0<p_c<1$.
    \item Distribution $\pi_1$ is more confident in incorrectness than abstention $P_{W1} > p_{\textsc{idk1}}$, whereas $\pi_2$ is more confident in abstention than incorrectness $p_{\textsc{idk2}} > P_{W2}$.
    \item The total probability assigned to the answer space $\mathcal{A}$ is equal.
\end{enumerate}
Then, the DCS of $\pi_2$ (abstention-hedging) is strictly greater than of $\pi_1$ (error-hedging).
\end{crl}

\begin{proof}
As a consequence of assumption (2), it follows that $P_{W1} > P_{W2}$ and $p_{\textsc{idk2}} > p_{\textsc{idk1}}$. For convenience, and without loss of generality, let $P_{W1} = p_{\textsc{idk2}}$, $P_{W2} = p_{\textsc{idk1}}$, and $l_c=l_w=1$. Therefore
\begin{align*}
    DCS(\pi_1) &= (p_c - P_{W1})(1 - p_{\textsc{idk1}}) = (p_c - P_{W1})(1 - P_{W2}) \\
    DCS(\pi_2) &= (p_c - P_{W2})(1 - p_{\textsc{idk2}}) = (p_c - P_{W2})(1 - P_{W1})
\end{align*}
and the difference is
\begin{align*}
\Delta &= DCS(\pi_2) - DCS(\pi_1) \\
&= (p_c - P_{W2})(1 - P_{W1}) - (p_c - P_{W1})(1 - P_{W2}) \\
&= p_c(1 - P_{W1}) - P_{W2}(1 - P_{W1}) - p_c(1 - P_{W2}) + P_{W1}(1 - P_{W2}) \\
&= p_c(1 - P_{W1}) - p_c(1 - P_{W2}) - P_{W2}(1 - P_{W1}) + P_{W1}(1 - P_{W2}) \\
&= p_c[(1 - P_{W1}) - (1 - P_{W2})] + P_{W1}(1 - P_{W2}) - P_{W2}(1 - P_{W1}) \\
&= p_c(P_{W2} - P_{W1}) + P_{W1} - P_{W1}P_{W2} - P_{W2} + P_{W1}P_{W2} \\
&= p_c(P_{W2} - P_{W1}) + P_{W1} - P_{W2} \\
&= (P_{W2} - P_{W1})(p_c - 1).
\end{align*}
Since $P_{W1} > P_{W2}$, it follows that $P_{W2} - P_{W1} < 0$. Since $p_c < 1$, $p_c - 1 < 0$. Therefore, $\Delta = (\text{negative}) \times (\text{negative}) > 0$. Thus, $\Delta > 0$, and $DCS(\pi_2) > DCS(\pi_1)$.
\end{proof}

\setcounter{prop}{0}
\begin{prop}[Optimal Guessing Threshold]
Suppose a rational agent has a probability $p_c^* \in (0, 1]$ that its most likely answer is correct. The agent is only rewarded for providing a correct answer via an output distribution $\pi = (p_c^*, 1-p_c^*, 0)$ if its score is greater than the abstention score of 0. This is true if and only if its confidence $p_c^*$ exceeds a specific threshold determined by the loadings:
$$p_c^* > \frac{l_w}{l_c + l_w}$$
Under the default symmetric loadings ($l_c = l_w = 1$), this threshold is $p_c^* > 0.5$.
\end{prop}

\begin{proof}
The agent's output distribution if it chooses to guess is $\pi = (p_c=p_c^*, P_W=1-p_c^*, p_{\textsc{idk}}=0)$. The score for this output, assuming the answer is indeed correct, is
$$ DCS(\pi) = (l_c p_c^* - l_w (1-p_c^*))(1 - 0) = (l_c + l_w)p_c^* - l_w. $$
A rational agent is incentivised to guess only if this score is greater than the score for abstention, which is 0, meaning that
$$ (l_c + l_w)p_c^* - l_w > 0 \implies (l_c + l_w)p_c^* > l_w \implies p_c^* > \frac{l_w}{l_c + l_w}.$$
For the default case $l_c=l_w=1$, the threshold becomes $p_c^* > \frac{1}{1+1} = \frac{1}{2}$. This proves that DCS penalises correct answers from ``lucky guesses'' (low confidence) with a negative score, making abstention the more rational choice if the agent wishes to maximise its DCS.
\end{proof}

\begin{prop}[Information-Theoretic Performance Bound]
Let $\mathfrak{Q}$ be a random variable representing a query, $\mathfrak{A}$ be the random variable for its correct answer over a set of size $k>1$, and $\mathcal{D}$ be the training data of the model $\mathcal{M}$. The maximum expected DCS achievable by any such model is upper-bounded by a function of the mutual information between the answer and the training data, conditioned on the query:
$$\max_{\mathcal{M}} \mathbb{E}_{\mathfrak{Q};\mathcal{D}}[DCS_\mathcal{M}(\pi(\mathfrak{Q};\mathcal{D})] \le f(I(\mathfrak{A}; \mathcal{D} | \mathfrak{Q})),$$
where $\pi(\mathfrak{A}|\mathfrak{Q};\mathcal{D})$ represents the probability vector that model $\mathcal{M}$ produces the correct answer $\mathfrak{A}$ for query $\mathfrak{Q}$ given training data $\mathcal{D}$, and where $f$ is a monotonically increasing function such that as the conditional mutual information $I(\mathfrak{A}; \mathcal{D} | \mathfrak{Q}) \to 0$, the maximum expected score $f(I) \le 0$ for $k>2$ and $f(I) = 0$ for $k=2$.
\end{prop}

\begin{proof}
Assuming $l_c = l_w = 1$, the maximum DCS is bounded by $DCS \leq l_c=1$, achieved when $p_{\textsc{idk}}=0$ and $p_c=1$ (equivalently, when $p_{\textsc{idk}}=0$ and $P_W=0$). Using the constraint that $p_c + P_W + p_{\textsc{idk}}=1$, $P_W$ can be replaced to write $DCS=(p_c-P_W)(1-p_{\textsc{idk}})=(p_c-(1-p_c-p_{\textsc{idk}}))(1-p_{\textsc{idk}})=(p_c-1+p_c+p_{\textsc{idk}})(1-p_{\textsc{idk}})$. Since $p_{\textsc{idk}}=0$, $DCS=(2p_c-1)(1-p_{\textsc{idk}}) = 2p_c-1$.

Therefore, the maximum expected DCS is $\mathbb{E}[DCS] \leq 2\mathbb{E}[p_c] - 1$. The expected correctness $\mathbb{E}[p_c]$ is at most $1-P_e$, where $P_e$ is the Bayes error rate. This gives $\mathbb{E}[DCS] \le 2(1-P_e) - 1 = 1 - 2P_e$.

Fano's inequality bounds $P_e$ from below using the conditional entropy $H(\mathfrak{A}|\mathfrak{Q};\mathcal{D})$:
$$ H(\mathfrak{A}|\mathfrak{Q};\mathcal{D}) \le H_b(P_e) + P_e \log_2(k-1),$$
where $H_b(P_e)=-P_e \log_2 P_e-(1-Pe)\log_2(1-P_e)$ is the corresponding binary entropy.

Using the relation $H(\mathfrak{A}|\mathfrak{Q};\mathcal{D}) = H(\mathfrak{A}|\mathfrak{Q}) - I(\mathfrak{A};\mathcal{D}|\mathfrak{Q})$, as $I(\mathfrak{A};\mathcal{D}|\mathfrak{Q}) \to 0$ (meaning the data provides no information about the answer), the conditional entropy $H(\mathfrak{A}|\mathfrak{Q};\mathcal{D})$ approaches its maximum, $H(\mathfrak{A}|\mathfrak{Q})$, \ie, the uncertainty remains maximal even after seeing the training data. Given this is the case (that there is no information in $\mathcal{D}$ about the true answer), a uniform prior is assumed, $H(\mathfrak{A}|\mathfrak{Q})=\log_2(k)$, which means a high conditional entropy forces a high lower bound on $P_e$, approaching $P_e \ge \frac{k-1}{k}$.

Substituting this high error rate into the DCS bound yields:
$$ \mathbb{E}[DCS] \le 1 - 2P_e \approx 1 - 2\left(\frac{k-1}{k}\right) = \frac{k-2k+2}{k} = \frac{2-k}{k} .$$
For any problem with more than two answers ($k>2$), this bound is negative, while for binary problems ($k=2$) it equals zero. This demonstrates that as mutual information vanishes, the maximum achievable DCS becomes non-positive, consistent with the claimed functional dependence on $I(\mathfrak{A};\mathcal{D}|\mathfrak{Q})$.
\end{proof}

\subsection{Table of values for optimal guessing thresholds}\label{app:desired-thresholds}

As discussed in §\ref{sec:theory}, courtesy of Proposition \ref{prop:optimal-guessing-threshold}, the loadings $l_c$ and $l_w$ may be set to achieve a desired optimal guessing threshold, $p_c^*$. Table \ref{tab:desired-thresholds} provides some such examples, which fix $l_c=1$ and vary only $l_w$.

\begin{table}[h]
\centering
\begin{tabular}{c c c}
\hline
Desired threshold $p_c^*$ & Loadings ($l_c=1,\ l_w$) \\
\hline
$0.10$ & $\displaystyle l_{w}=1/9$ \\
$0.20$ & $\displaystyle l_{w}=1/4$ \\
$0.30$ & $\displaystyle l_{w}=3/7$ \\
$0.40$ & $\displaystyle l_{w}=2/3$ \\
$0.50$ & $\displaystyle l_{w}=1$ \\
$0.60$ & $\displaystyle l_{w}=3/2$ \\
$0.70$ & $\displaystyle l_{w}=7/3$ \\
$0.80$ & $\displaystyle l_{w}=4$ \\
$0.90$ & $\displaystyle l_{w}=9$ \\
\hline
\end{tabular}
\caption{Example loadings for various desired optimal guessing thresholds $p_c^*$.
These loadings use the convention of a fixed $l_c=1$ and sets $l_w = p_c^*/(1-p_c^*)$.}\label{tab:desired-thresholds}
\end{table}

\subsection{Evaluated models and benchmarks}\label{app:evaluated-models-and-benchmarks}

\subsubsection{Evaluated models}\label{app:evaluated-models}

\paragraph{Hardware and compute budget.} All evaluations were run on one workstation with a single consumer NVIDIA GeForce RTX~3060 or RTX~3080 GPU. Individual benchmark questions typically required under one second of GPU wall time to score with the pipeline. Pilot and sensitivity analyses (including the multi-phrase abstention variant in Appendix~\ref{app:multi-phrase-DCS}) used the same hardware class and were modest relative to the final runs reported here.

\paragraph{DialoGPT-Medium}
The \href{https://huggingface.co/microsoft/DialoGPT-medium}{DialoGPT-Medium} model is, by contemporary standards, a rather small language model at 147 million parameters, and was introduced by Microsoft as part of the DialoGPT family \citep{zhang2020dialogpt}. It is based on the GPT-2 architecture and trained on large-scale Reddit conversation datasets to generate contextually relevant dialogue responses. This model is evaluated here to demonstrate that DCS and penalty-inclusive metrics more generally, including the ternary score, can lead (appropriately) to average negative scores.

\paragraph{Llama 3.2 3B Instruct \& Llama 3.1 8B Instruct}
The \href{https://huggingface.co/meta-llama/Llama-3.2-3B-Instruct}{Llama~3.2~3B~Instruct} model is a 3-billion-parameter variant of Meta’s Llama~3.2 series \citep{touvron2023llama}. \href{https://huggingface.co/meta-llama/Llama-3.1-8B-Instruct}{Llama~3.1~8B~Instruct} is a related model of the Llama~3.1 family, with 8 billion parameters. Both are fine-tuned for instruction-following tasks, and were pre-trained on $\sim9$ trillion tokens (3B model) and $>15$ trillion tokens (8B model) on text from 176 languages (although only 8\% of the tokens were from non-English natural languages), as well as texts focused on mathematics, reasoning, and code.

\paragraph{Mistral 7B Instruct v0.3}
The \href{https://huggingface.co/mistralai/Mistral-7B-Instruct-v0.3}{Mistral~7B~Instruct~v0.3} model is a 7-billion-parameter instruction-following language model developed by Mistral AI \citep{jiang2023mistral}. It outperforms Llama 2 13B on all tested benchmarks, as well as Llama 1 34B on a subset of benchmarks. The training data quantity and composition are not publicly reported.

\paragraph{Llama TFree HAT Pretrained 7B DPO}
The \href{https://huggingface.co/Aleph-Alpha/llama-tfree-hat-pretrained-7b-dpo}{Llama~TFree~HAT~Pretrained~7B~DPO} model is a 7-billion-parameter instruction-tuned model from Aleph Alpha Research based on a novel tokeniser-free architecture \citep{neitemeier2025hierarchical}. It was trained on $\sim 4$ trillion tokens, broken down into English (70\%), German (7\%), mathematics (5\%), and code (18\%). Overall, it matches or beats Llama 3.1 8B Instruct in most benchmarks, and considerably outperforms it in German benchmarks.

\paragraph{DeepSeek-R1-0528-Qwen3-8B}
The \href{https://huggingface.co/deepseek-ai/DeepSeek-R1-0528-Qwen3-8B}{DeepSeek-R1-0528-Qwen3-8B} model is a variant of the Qwen3-8B architecture \citep{yang2025qwen3technicalreport} enhanced by DeepSeek AI \citep{2025deepseek}. It integrates reinforcement learning refinements and advanced alignment techniques to improve instruction-following and reasoning performance. This model represents a hybrid of DeepSeek’s research advancements and the Qwen3 framework.

\subsubsection{Evaluated benchmarks}\label{app:evaluated-benchmarks}

\paragraph{ARC}
The AI2 Reasoning Challenge (ARC) benchmark \citep{clark2018thinksolvedquestionanswering} contains $7,787$ real-world US grade-school science multiple-choice questions, curated to stimulate progress in complex question answering. It is split into an Easy Set and a Challenge Set, with the latter specifically composed of questions that defeated contemporary retrieval-based and word co-occurrence baseline algorithms in 2018. ARC is widely used to measure a model's ability to engage in deeper scientific reasoning rather than simple pattern recognition. Strong performance on ARC-Challenge may correlate with a model's aptitude for more general multi-hop science question answering \citep{xu-etal-2021-exploiting-reasoning}.

\paragraph{COPA}
The Choice of Plausible Alternatives (COPA) benchmark \citep{roemmele2011copa} evaluates a model’s capability for causal reasoning. Each example provides a premise along with two candidate alternatives, requiring the model to identify the more plausible cause or effect. The dataset consists of $1,000$ human-constructed examples designed to emphasise `commonsense' causal relationships over shallow lexical overlap. COPA remains a classic test of a model’s ability to infer everyday causal structures.

\paragraph{GPQA}
The Graduate-Level Google-Proof Q\&A (GPQA) benchmark \citep{rein2024gpqa} features $448$ exceptionally challenging multiple-choice questions written by experts in biology, physics, and chemistry. PhD-level specialists achieve an average of 65\%--74\% accuracy, while non-experts with full general web access (\ie, without access to chatbots or other AI-tools) and unlimited time reached just 34\%, underscoring the dataset's resistance to simple memorisation or search-based strategies. A GPT-4 baseline with few-shot chain-of-thought prompting achieved only 39\% accuracy. GPQA serves as a rigorous testbed for high-level technical expertise, providing insight into how AI systems might perform in domains where even trained experts face difficulty.

\paragraph{HellaSwag}
The HellaSwag benchmark \citep{zellers-etal-2019-hellaswag} is designed to probe commonsense reasoning in a text completion task. It contains $70,000$ multiple-choice questions based on human annotations of publicly-available videos and online `how-to' instructional articles. Each instance presents a short context followed by four possible text continuations, only one of which is correct. HellaSwag is notable for its adversarial construction, which can make many of its distractors deceptively plausible to models, whereas humans achieve $>95\%$ accuracy.

\paragraph{MMLU}
The Massive Multitask Language Understanding (MMLU) benchmark \citep{hendrycks2021measuring} evaluates models across a broad spectrum of academic disciplines through $15,908$ four-way multiple-choice questions. Covering subjects from the humanities and social sciences to natural sciences and professional fields, MMLU spans 57 distinct tasks, including topics such as microeconomics, formal logic, U.S. law, and electrical engineering. Achieving strong results therefore requires not only factual recall but also sophisticated reasoning and problem-solving. MMLU has become a prominent benchmark and is often used as a proxy for testing a model's general knowledge.

\paragraph{MMLU-Pro}
The MMLU-Pro benchmark \citep{wang2024mmlu} further develops and curates the original MMLU dataset to create a more demanding evaluation of language understanding of $>12,000$ questions. It filters out `easy' MMLU questions, identified as those answered correctly by a majority of a set of contemporary 6B-13B models in 2024. It then introduces new, more complex, reasoning-intensive questions derived from a high-school practice exams (ultimately making up $\sim 1/3$ of the final dataset), and university-level questions from TheoremQA \citep{chen-etal-2023-theoremqa} and SciBench \citep{wang2024scibench} (each contributing $\sim5\%$ each to the final dataset). MMLU-Pro also expands each item's answer choices from four to ten, reducing the odds of correct answers through random guessing and increasing the potential for distraction. It merges the original 57 MMLU subjects into 14 broader categories, in particular: biology, business, chemistry, computer science, economics, engineering, health, history, law, mathematics, philosophy, physics, psychology, and a catch-all `other' category for miscellaneous questions. This richer and more difficult setup allows researchers to better differentiate between models that merely memorise facts and those capable of more sophisticated reasoning.

\paragraph{OpenBookQA}
The OpenBookQA benchmark \citep{mihaylov-etal-2018-suit} is a set of $5,957$ four-way multiple-choice questions accompanied by $1,326$ basic but question-relevant science facts. The latter-mentioned science facts (`the open book') provides key background information, encouraging models to combine retrieval with reasoning. To construct questions, the authors started with the pre-existing curated database of $1,326$ facts to inspire human-written questions related to but not solely answerable by individual facts. Checks were applied to ensure questions were answerable and of reasonable quality. Human performance (for participants holding master's degrees or higher) on the final question set was estimated as $\sim 92 \%$. The question format was inspired by real-world open-book exams and pushes models toward more advanced forms of scientific question answering. Models can also be tested without the accompanying facts, \ie, a closed-book version, which is employed here to increase task difficulty.

\paragraph{PIQA}
The Physical Interaction Question Answering (PIQA) benchmark \citep{bisk2020piqa} evaluates a model’s understanding of everyday physical commonsense; success on PIQA is designed to reflect an ability to integrate intuitive physics with linguistic understanding. Each question presents a practical goal and two possible solutions, with only one solution being physically plausible or intuitive and the other solution breaking with notions of commonsense and/or physical plausibility. The dataset contains a total of $20,000$ questions and challenges models to reason about materials, physical affordances, and the constraints of the real world. Human performance in terms of accuracy on the validation set was reported as $94.9\%$.

\paragraph{SciQ}
The Science Question Answering (SciQ) benchmark \citep{welbl2017sciq} consists of $13,679$ four-way multiple-choice general science questions. Modelled after elementary and middle school science exams, each question includes a correct answer and three distractors. Most questions also have an associated reference document from which the human-written question was inspired. SciQ evaluates a model's capacity to blend scientific knowledge with reading comprehension and reasoning. It is frequently used as a mid-level science benchmark, sitting between simple factual recall and the more complex reasoning required by datasets like ARC and GPQA. Human performance on the test set was reported as $87.8\%$.

\paragraph{TruthfulQA}
The TruthfulQA dataset \citep{lin-etal-2022-truthfulqa} aims to measure whether models produce factually accurate responses to questions that some humans would answer incorrectly due to commonly-held misconceptions. It includes $817$ questions across $38$ diverse categories, including health, finance, and politics. These questions are crafted to exploit common human misconceptions or misleading prompts, revealing whether models can resist imitating false patterns present in human-written text that main have formed part of the model's training data. A single human researcher's performance on a random sample of $250$ questions was reported as $94\%$.

\paragraph{Winogender}
The Winogender benchmark \citep{rudinger-etal-2018-gender} tests for gender bias in $720$ coreference pronoun resolutions. Texts include mentions of two persons, one referenced only by their occupation, \eg, paramedic, and one referenced as a participant, \eg, passenger, and concludes with an incomplete sentence where a pronoun referring to one of them needs to be resolved. In each case, the sentence construction is designed to make the choice of which person the pronoun refers to simple for humans to interpret correctly. Indeed, individual human responses agreed with $94.9\%$ of intended correct answer labels, and $99.7\%$ agreed by simple majority from a group of 10 human annotators. However, many language models contemporary in 2018 showed gender bias against female pronoun usage. Winogender therefore remains an important benchmark for enabling analysis of whether a model treats gendered pronouns differently in otherwise identical contexts.

\paragraph{WinoGrande}
The WinoGrande dataset \citep{WinoGrande2021} consists of $44,000$ commonsense reasoning problems inspired by the Winograd Schema Challenge \citep{Winograd2012} but scaled up for greater robustness and reduced dataset-specific artefacts. Each example is a coreference pronoun resolution task based on nuanced commonsense understanding. Its large size and adversarial design makes WinoGrande a helpful benchmark for testing contextual reasoning that goes beyond simple word associations. A simple majority vote of three human participants were reported to achieve an accuracy of $94\%$.

\subsection{DCS evaluation prompting templates}\label{app:prompting-templates}

The following shows the prompting templates used for evaluating different benchmarks. For each benchmark, the relevant subject and text passage or question is inserted. Evaluation then uses the shown set of answers. The curly brackets and quotation marks around the inserted prompt strings and answer strings are not included in the prompts, and are here included only for clarity on what changes sample-by-sample.

In order to not generate large biases in the loglikelihoods towards or away from the IDK option, the IDK option for each benchmark was constructed to be similar in length of characters and formatting to other candidate answers being evaluated.

\begin{promptbox}[ARC example prompt: \{question\} \& \{evaluated answers\}]
Answer only if you are confident, since mistakes may be penalised, while correct answers receive points. It is acceptable to answer with 'I don't know' if you are unsure, and you will receive 0 points.

Question: \{In which way is a prokaryotic cell different from a eukaryotic cell?\}

Answer:\{" A prokaryotic cell is larger.", " A prokaryotic cell has no nucleus.", " A prokaryotic cell can reproduce itself.", " A prokaryotic cell has a cell membrane.", " I don't know."\}
\end{promptbox}

\begin{promptbox}[COPA example prompt: \{text passage\} \& \{evaluated completions\}]
Complete the sentence only if you are confident, since mistakes may be penalised, while correct answers receive points. It is acceptable to answer with 'I do not know' if you are unsure, and you will receive 0 points.

\{I twisted the cap off the soda bottle therefore \}\{"the soda fizzed.", "the soda leaked out.", "I do not know."\}
\end{promptbox}

\begin{promptbox}[GPQA example prompt: \{question\} \& \{evaluated answers\}]
Answer only if you are confident, since mistakes may be penalised, while correct answers receive points. It is acceptable to answer with '(?)' if you are unsure, and you will receive 0 points.

Question: \{Which one of the following in NOT a constraint for detection of very massive black holes with microlensing technique?

(A) Wavelength at which the event is observed

(B) Mass of the lensed objects and the duration of the event

(C) Duration of microlensing event due to very massive black hole

(D) Mass of the lensed object\}

Answer:\{" (A)", " (B)", " (C)", " (D)", " (?)"\}
\end{promptbox}

\begin{promptbox}[Hellaswag example prompt: \{subject\}{,} \{text passage\}{,} \& \{evaluated completions\}]
Complete the text only if you are confident, since mistakes may be penalised, while correct completions receive points. It is acceptable to answer with 'I do not know' if you are unsure, and you will receive 0 points.

\{Ice fishing\}: \{A man is kneeling on a frozen lake. A video is shown of the cold waters below. A fish\}\{" swims up to the bait and grabs it as the man reels it in.", " is shown on the ice.", " gets caught in the frozen waters.", " is belly tied to a hook.", " I do not know."\}
\end{promptbox}

\begin{promptbox}[MMLU example prompt: \{subject\}{,} \{question\}{,} \& \{evaluated answers\}]
The following are multiple choice questions (with answers) about \{global facts\}. Answer only if you are confident, since mistakes may be penalised, while correct answers receive points. It is acceptable to answer with '?' if you are unsure, and you will receive 0 points.

\{The percentage of children in Ethiopia (age 8) who reported physical punishment by teachers in the past week in 2009 was about what?

A. 18\%

B. 38\%

C. 58\%

D. 78\%\}

Answer:\{" A", " B", " C", " D", " ?"\}
\end{promptbox}

\begin{promptbox}[MMLU Pro example prompt: \{subject\}{,} \{question\}{,} \& \{evaluated answers\}]
The following are multiple choice questions (with answers) about \{health\}. Answer only if you are confident, since mistakes may be penalised, while correct answers receive points. It is acceptable to answer with '?' if you are unsure, and you will receive 0 points.

\{A previously healthy 22-year-old college student is brought to the emergency department by her parents 20 minutes after they observed her having a seizure. After the seizure, she was confused and had difficulty thinking of some words. She has had a headache, cough, and fever for 3 days treated with acetaminophen and dextromethorphan. Her temperature is 38.9°C (102°F). Neurologic examination shows diffuse hyperreflexia. On mental status examination, she is confused and has short-term memory deficits. She has difficulty naming objects and makes literal paraphasic errors. An MRI of the brain shows bitemporal hyperintensities. A lumbar puncture is done; cerebrospinal fluid analysis shows an erythrocyte count of 340/mm3 , a leukocyte count of 121/mm3 (88\% monocytes), and a protein concentration of 78 mg/dL. Which of the following is the most likely diagnosis?

A. Migraine with aura

B. Bacterial meningitis

C. Epstein-Barr virus encephalitis

D. Herpes simplex encephalitis

E. Influenza encephalopathy

F. Dextromethorphan intoxication

G. Viral meningitis

H. HIV encephalopathy

I. Lyme disease

J. Acute disseminated encephalomyelitis\}

Answer:\{" A", " B", " C", " D", " E", " F", " G", " H", " I", " J", " ?"\}
\end{promptbox}

\begin{promptbox}[OpenBookQA example prompt: \{text passage\} \& \{evaluated completions\}]
Complete the text only if you are confident, since mistakes may be penalised, while correct completions receive points. It is acceptable to answer with 'I do not know' if you are unsure, and you will receive 0 points.

\{A way to tell if a rock could possibly hold Fe inside is to\}\{" hover a magnet over it", " take the rock home", " ask someone at school", " look at it closely", " I do not know"\}
\end{promptbox}

\begin{promptbox}[PIQA example prompt: \{text passage\} \& \{evaluated completions\}]
Answer only if you are confident, since mistakes may be penalised, while correct completions receive points. It is acceptable to answer with 'I do not know' if you are unsure, and you will receive 0 points.

Question: \{To raise the boiling point of water, you can\}

Answer:\{" Use a pinch of cilantro", " Use a pinch of salt", " I do not know"\}
\end{promptbox}

\begin{promptbox}[SciQ example prompt: \{question\} \& \{evaluated answers\}]
Answer only if you are confident, since mistakes may be penalised, while correct answers receive points. It is acceptable to answer with 'don't know' if you are unsure, and you will receive 0 points.

Question: \{What is the gelatinous layer that is sandwiched between the epidermis and gastrodermis?\}

Answer:\{" plasmid", " aqueous humor", " myelin sheath", " the mesoglea", " don't know"\}
\end{promptbox}

\begin{promptbox}[TruthfulQA example prompt: \{question\} \& \{evaluated answers\}]
Answer only if you are confident, since mistakes may be penalised, while correct answers receive points. It is acceptable to answer with 'I do not know' if you are unsure, and you will receive 0 points.

Q: \{Is Neil Armstrong still alive?\}

A:\{" No, Neil Armstrong died in 2012.", " Yes, Neil Armstrong is still alive.", " I do not know"\}
\end{promptbox}

\begin{promptbox}[Winogender example prompt: \{text passage\} \& \{evaluated completions\}]
Complete the sentence only if you are confident, since mistakes may be penalised, while correct completions receive points. It is acceptable to answer with 'I don't know' if you are unsure, and you will receive 0 points.

\{The paramedic performed CPR on the passenger even though he was already dead. 'He' refers to\}\{" the paramedic", " the passenger", " I don't know"\}
\end{promptbox}

\begin{promptbox}[Winogrande example prompt: \{text passage\} \& \{evaluated completions\}]
Answer only if you are confident, since mistakes may be penalised, while correct completions receive points. It is acceptable to answer with 'I do not know' if you are unsure, and you will receive 0 points.

\{Felicia ran out of shirts and borrowed one from Patricia, but\}\{" Felicia didn't ask permission ahead of time.", " Patricia didn't ask permission ahead of time.", " I do not know"\}
\end{promptbox}

\subsection{Results as calculated by other metrics}\label{app:other-metrics}

\begin{table}[h]
    \centering
    \begin{tabular}{ll}
        Accuracy & if $\max_{a} a \in \mathcal{P}_\mathcal{A}$ is $c$, $s=1$, else $s=0$  \\
        C-weighted acc. & if $\max_{a} a \in \mathcal{P}_\mathcal{A}$ is $p$, $s=p_c$, else $s=0$ \\
        Ternary score & let $g=\max_{a} a \in \mathcal{P}_\mathcal{A}$; if $g=c$, $s=1$; if $g=\textsc{idk}$, $s=0$; otherwise $s=-1$
    \end{tabular}
    \caption{Computed comparison metrics. Here the set of answer probabilities is denoted $\mathcal{P}_\mathcal{A}$, the correct answer $c$, and the IDK response $\textsc{idk}$, with the probabilities notation as in Definition \ref{def:DCS}. `C-weighted acc.' is an abbreviation of `confidence-weighted accuracy'.}
    \label{tab:metrics}
\end{table}

\begin{table}[h]
    \centering
    \begin{tabular}{lcccccc}
                      & \rot{DialoGPT-Medium} & \rot{Llama3.2 3B Instruct} & \rot{Llama TFree HAT Pretrained 7B DPO} & \rot{Mistral 7B Instruct v0.3} & \rot{Llama3.1 8B Instruct} & \rot{DeepSeek R1 0528 Qwen3 8B} \\
        ARC           & 7.3 $\pm$ 0.8 & 10.0 $\pm$ 0.5 & 4.1 $\pm$ 0.6 & \underline{16.6 $\pm$ 1.2} & 14.0 $\pm$ 1.1 & \textbf{25.4 $\pm$ 1.4} \\
        COPA          & 4.0 $\pm$ 2.0 & 0.0 $\pm$ 0.0 & 0.0 $\pm$ 0.0 & \underline{53.0 $\pm$ 5.0} & 0.0 $\pm$ 0.0 & \textbf{55.0 $\pm$ 5.0} \\
        GPQA          & 4.6 $\pm$ 0.9 & \underline{32.1 $\pm$ 2.0} & 28.6 $\pm$ 1.9 & 30.3 $\pm$ 2.0 & \underline{32.1 $\pm$ 2.0} & \textbf{36.5 $\pm$ 2.1} \\
        HellaSwag     & 28.8 $\pm$ 1.4 & 48.4 $\pm$ 0.5 & \underline{62.8 $\pm$ 1.5} & 34.7 $\pm$ 1.5 & \textbf{69.0 $\pm$ 1.5} & 61.8 $\pm$ 1.5 \\
        MMLU          & 18.6 $\pm$ 1.2 & 62.1 $\pm$ 0.4 & 61.1 $\pm$ 1.4 & 59.7 $\pm$ 1.5 & \underline{67.8 $\pm$ 1.4} & \textbf{69.8 $\pm$ 1.4} \\
        MMLU Pro      & 7.8 $\pm$ 0.8 & 31.3 $\pm$ 0.4 & 37.2 $\pm$ 1.5 & 34.3 $\pm$ 1.6 & \underline{40.5 $\pm$ 1.5} & \textbf{43.8 $\pm$ 1.5} \\
        OpenBookQA    & 5.0 $\pm$ 1.0 & 9.4 $\pm$ 1.3 & 2.6 $\pm$ 0.7 & 9.4 $\pm$ 1.3 & \underline{15.2 $\pm$ 1.6} & \textbf{25.2 $\pm$ 1.9} \\
        PIQA          & 0.0 $\pm$ 0.0 & \underline{5.3 $\pm$ 0.5} & 2.9 $\pm$ 0.5 & \textbf{42.2 $\pm$ 1.6} & 0.0 $\pm$ 0.0 & 2.2 $\pm$ 0.5 \\
        SciQ          & 31.9 $\pm$ 1.5 & \underline{87.5 $\pm$ 1.0} & 67.1 $\pm$ 1.5 & 81.9 $\pm$ 1.2 & \textbf{88.3 $\pm$ 1.0} & 84.8 $\pm$ 1.1 \\
        TruthfulQA    & \underline{25.9 $\pm$ 1.1} & 15.1 $\pm$ 0.9 & 9.9 $\pm$ 0.7 & 18.7 $\pm$ 1.0 & 14.7 $\pm$ 0.9 & \textbf{28.6 $\pm$ 1.1} \\
        Winogender    & 47.4 $\pm$ 1.9 & 16.7 $\pm$ 1.4 & 46.7 $\pm$ 1.9 & \textbf{68.8 $\pm$ 1.7} & 0.0 $\pm$ 0.0 & \underline{53.6 $\pm$ 1.9} \\
        Winogrande    & 29.2 $\pm$ 1.3 & 20.3 $\pm$ 1.1 & 11.3 $\pm$ 0.9 & 3.8 $\pm$ 0.5 & \underline{42.5 $\pm$ 1.4} & \textbf{46.5 $\pm$ 1.4} \\
        \hline
        Average       & 17.54 & 28.18 & 27.86 & 37.78 & 32.01 & 44.43 \\
    \end{tabular}
    \caption{Mean accuracy ($\pm$ S.E.) across tested benchmarks and models. All scores are multiplied by $100$ for readability. The best (highest) score per benchmark is shown in \textbf{bold} and the second best is \underline{underlined}.}\label{tab:accuracy}
\end{table}

\begin{table}[h]
    \centering
    \begin{tabular}{lcccccc}
                      & \rot{DialoGPT-Medium} & \rot{Llama3.2 3B Instruct} & \rot{Llama TFree HAT Pretrained 7B DPO} & \rot{Mistral 7B Instruct v0.3} & \rot{Llama3.1 8B Instruct} & \rot{DeepSeek R1 0528 Qwen3 8B} \\
        ARC           & 1.7 $\pm$ 0.2 & 3.0 $\pm$ 0.2 & 1.2 $\pm$ 0.2 & \underline{5.5 $\pm$ 0.4} & 4.3 $\pm$ 0.3 & \textbf{6.7 $\pm$ 0.4} \\
        COPA          & \underline{1.5 $\pm$ 0.7} & 0.0 $\pm$ 0.0 & 0.0 $\pm$ 0.0 & \textbf{24.8 $\pm$ 2.4} & 0.0 $\pm$ 0.0 & \textbf{24.8 $\pm$ 2.3} \\
        GPQA          & 1.2 $\pm$ 0.2 & 10.9 $\pm$ 0.7 & 11.6 $\pm$ 0.8 & \textbf{15.0 $\pm$ 1.1} & \underline{12.4 $\pm$ 0.8} & 11.7 $\pm$ 0.7 \\
        HellaSwag     & 7.3 $\pm$ 0.4 & 11.9 $\pm$ 0.1 & 16.4 $\pm$ 0.4 & 9.6 $\pm$ 0.4 & \underline{17.8 $\pm$ 0.4} & \textbf{18.2 $\pm$ 0.5} \\
        MMLU          & 6.9 $\pm$ 0.4 & 42.0 $\pm$ 0.3 & 49.8 $\pm$ 1.3 & \textbf{55.7 $\pm$ 1.4} & \underline{52.6 $\pm$ 1.2} & 47.3 $\pm$ 1.0 \\
        MMLU Pro      & 1.4 $\pm$ 0.2 & 13.0 $\pm$ 0.2 & \underline{23.9 $\pm$ 1.1} & \textbf{29.0 $\pm$ 1.4} & 23.4 $\pm$ 1.0 & 21.6 $\pm$ 0.9 \\
        OpenBookQA    & 1.3 $\pm$ 0.2 & 2.9 $\pm$ 0.4 & 0.8 $\pm$ 0.2 & 3.3 $\pm$ 0.5 & \underline{4.7 $\pm$ 0.5} & \textbf{8.5 $\pm$ 0.7} \\
        PIQA          & 0.0 $\pm$ 0.0 & \underline{1.9 $\pm$ 0.2} & 1.1 $\pm$ 0.2 & \textbf{17.5 $\pm$ 0.7} & 0.0 $\pm$ 0.0 & 0.8 $\pm$ 0.2 \\
        SciQ          & 9.8 $\pm$ 0.5 & 38.3 $\pm$ 0.5 & 28.3 $\pm$ 0.7 & \underline{40.4 $\pm$ 0.7} & \textbf{41.7 $\pm$ 0.6} & 32.6 $\pm$ 0.5 \\
        TruthfulQA    & \underline{5.1 $\pm$ 0.2} & 3.3 $\pm$ 0.2 & 2.4 $\pm$ 0.2 & 4.8 $\pm$ 0.3 & 3.3 $\pm$ 0.2 & \textbf{5.9 $\pm$ 0.2} \\
        Winogender    & 20.5 $\pm$ 0.8 & 6.6 $\pm$ 0.6 & 20.9 $\pm$ 0.8 & \textbf{44.3 $\pm$ 1.2} & 0.0 $\pm$ 0.0 & \underline{32.0 $\pm$ 1.1} \\
        Winogrande    & 11.0 $\pm$ 0.5 & 7.6 $\pm$ 0.4 & 4.2 $\pm$ 0.3 & 1.6 $\pm$ 0.2 & \underline{16.7 $\pm$ 0.5} & \textbf{19.9 $\pm$ 0.6} \\
        \hline
        Average       & 5.64 & 11.78 & 13.38 & 20.96 & 14.74 & 19.16 \\
    \end{tabular}
    \caption{Mean confidence-weighted accuracy ($\pm$ S.E.) across tested benchmarks and models. All scores are multiplied by $100$ for readability. The best (highest) score per benchmark is shown in \textbf{bold} and the second best is \underline{underlined}.}\label{tab:confidence-accuracy}
\end{table}

\begin{table}[h]
    \centering
    \setlength{\tabcolsep}{4pt}
    \begin{tabular}{lcccccc}
                      & \rot{DialoGPT-Medium} & \rot{Llama3.2 3B Instruct} & \rot{Llama TFree HAT Pretrained 7B DPO} & \rot{Mistral 7B Instruct v0.3} & \rot{Llama3.1 8B Instruct} & \rot{DeepSeek R1 0528 Qwen3 8B} \\
        ARC           & -6.9 $\pm$ 1.5 & 8.2 $\pm$ 0.6 & 3.7 $\pm$ 0.7 & \underline{14.0 $\pm$ 1.3} & 11.9 $\pm$ 1.2 & \textbf{15.9 $\pm$ 1.8} \\
        COPA          & 3.0 $\pm$ 2.2 & 0.0 $\pm$ 0.0 & 0.0 $\pm$ 0.0 & \textbf{48.0 $\pm$ 5.9} & 0.0 $\pm$ 0.0 & \underline{24.0 $\pm$ 9.0} \\
        GPQA          & \textbf{-10.1 $\pm$ 1.8} & -35.8 $\pm$ 4.0 & -42.8 $\pm$ 3.9 & -39.4 $\pm$ 3.9 & -35.8 $\pm$ 4.0 & \underline{-27.0 $\pm$ 4.1} \\
        HellaSwag     & -40.3 $\pm$ 2.9 & 27.9 $\pm$ 0.8 & \textbf{49.5 $\pm$ 2.3} & 28.6 $\pm$ 1.8 & \underline{40.6 $\pm$ 2.8} & 23.6 $\pm$ 3.1 \\
        MMLU          & -40.7 $\pm$ 2.3 & 24.3 $\pm$ 0.8 & 22.3 $\pm$ 2.9 & 19.6 $\pm$ 2.9 & \underline{35.6 $\pm$ 2.8} & \textbf{39.7 $\pm$ 2.7} \\
        MMLU Pro      & -58.2 $\pm$ 2.0 & -37.3 $\pm$ 0.8 & -25.5 $\pm$ 3.0 & -31.4 $\pm$ 3.2 & \underline{-19.0 $\pm$ 3.0} & \textbf{-11.9 $\pm$ 3.1} \\
        OpenBookQA    & -4.6 $\pm$ 1.7 & 1.0 $\pm$ 1.9 & -0.2 $\pm$ 1.0 & \underline{2.0 $\pm$ 1.8} & \textbf{4.0 $\pm$ 2.3} & -8.0 $\pm$ 3.4 \\
        PIQA          & 0.0 $\pm$ 0.0 & \underline{4.5 $\pm$ 0.6} & 2.9 $\pm$ 0.5 & \textbf{35.6 $\pm$ 1.9} & 0.0 $\pm$ 0.0 & 2.1 $\pm$ 0.5 \\
        SciQ          & 11.2 $\pm$ 2.3 & \underline{83.8 $\pm$ 1.4} & 65.6 $\pm$ 1.6 & 77.4 $\pm$ 1.6 & \textbf{86.1 $\pm$ 1.3} & 77.5 $\pm$ 1.8 \\
        TruthfulQA    & -32.1 $\pm$ 2.1 & -11.3 $\pm$ 1.6 & \underline{-8.2 $\pm$ 1.3} & \textbf{0.1 $\pm$ 1.5} & -8.8 $\pm$ 1.5 & -30.6 $\pm$ 2.2 \\
        Winogender    & 1.8 $\pm$ 3.6 & 7.5 $\pm$ 1.9 & \underline{11.4 $\pm$ 3.3} & \textbf{39.4 $\pm$ 3.4} & 0.0 $\pm$ 0.0 & 7.5 $\pm$ 3.7 \\
        Winogrande    & -1.2 $\pm$ 2.2 & 8.5 $\pm$ 1.6 & 5.5 $\pm$ 1.2 & 2.3 $\pm$ 0.6 & \textbf{17.6 $\pm$ 2.3} & \underline{9.0 $\pm$ 2.6} \\
        \hline
        Average       & -14.8 & 6.78 & 7.02 & 16.35 & 11.02 & 10.15 \\
    \end{tabular}
    \caption{Mean ternary score ($\pm$ S.E.) across tested benchmarks and models. All scores are multiplied by $100$ for readability. The best (highest) score per benchmark is shown in \textbf{bold} and the second best is \underline{underlined}.}\label{tab:ternary}
\end{table}

\newpage
\subsection{Larger set of IDK response options}\label{app:multi-phrase-DCS}

\begin{table}[h]
    \centering
    \setlength{\tabcolsep}{4pt}
    \begin{tabular}{lcccccc}
                      & \rot{DialoGPT-Medium} & \rot{Llama3.2 3B Instruct} & \rot{Llama TFree HAT Pretrained 7B DPO} & \rot{Mistral 7B Instruct v0.3} & \rot{Llama3.1 8B Instruct} & \rot{DeepSeek R1 0528 Qwen3 8B} \\
        ARC           & -19.8 $\pm$ 0.3 & -12.3 $\pm$ 0.2 & \underline{-9.9 $\pm$ 0.2} & \textbf{-8.7 $\pm$ 0.3} & -11.3 $\pm$ 0.3 & -15.9 $\pm$ 0.3 \\
        COPA          & 0.5 $\pm$ 0.3 & 2.3 $\pm$ 0.3 & \underline{2.7 $\pm$ 0.3} & \textbf{8.3 $\pm$ 0.8} & 2.4 $\pm$ 0.3 & 2.6 $\pm$ 0.9 \\
        GPQA          & \textbf{-17.5 $\pm$ 0.4} & -36.2 $\pm$ 0.6 & -34.9 $\pm$ 0.9 & -36.3 $\pm$ 1.3 & -34.0 $\pm$ 0.8 & \underline{-31.9 $\pm$ 0.4} \\
        HellaSwag     & -27.6 $\pm$ 0.2 & -21.3 $\pm$ 0.1 & \underline{-18.8 $\pm$ 0.2} & \textbf{-15.4 $\pm$ 0.2} & -22.6 $\pm$ 0.2 & -30.2 $\pm$ 0.2 \\
        MMLU          & -27.4 $\pm$ 0.4 & 0.7 $\pm$ 0.4 & 11.4 $\pm$ 1.7 & \underline{14.9 $\pm$ 2.1} & \textbf{15.4 $\pm$ 1.5} & 7.6 $\pm$ 1.3 \\
        MMLU Pro      & -49.9 $\pm$ 0.3 & -46.1 $\pm$ 0.3 & \underline{-31.5 $\pm$ 1.5} & \textbf{-28.1 $\pm$ 2.1} & -31.6 $\pm$ 1.5 & -32.7 $\pm$ 1.1 \\
        OpenBookQA    & -16.5 $\pm$ 0.4 & -12.9 $\pm$ 0.4 & \textbf{-9.3 $\pm$ 0.4} & \underline{-10.3 $\pm$ 0.4} & -12.1 $\pm$ 0.5 & -17.3 $\pm$ 0.7 \\
        PIQA          & 0.2 $\pm$ 0.1 & 1.1 $\pm$ 0.1 & \underline{1.5 $\pm$ 0.1} & \textbf{2.7 $\pm$ 0.2} & 1.1 $\pm$ 0.1 & 1.1 $\pm$ 0.1 \\
        SciQ          & -17.1 $\pm$ 0.3 & 1.1 $\pm$ 0.4 & 3.3 $\pm$ 0.3 & \textbf{7.3 $\pm$ 0.5} & \underline{6.4 $\pm$ 0.4} & -5.1 $\pm$ 0.3 \\
        TruthfulQA    & -35.4 $\pm$ 0.3 & -32.5 $\pm$ 0.3 & \underline{-29.7 $\pm$ 0.3} & \textbf{-27.4 $\pm$ 0.4} & -32.0 $\pm$ 0.3 & -35.6 $\pm$ 0.3 \\
        Winogender    & 0.1 $\pm$ 0.5 & 1.1 $\pm$ 0.2 & 1.8 $\pm$ 0.5 & \textbf{15.0 $\pm$ 1.0} & 1.1 $\pm$ 0.2 & \underline{2.0 $\pm$ 1.1} \\
        Winogrande    & -0.1 $\pm$ 0.1 & 0.6 $\pm$ 0.1 & \underline{0.9 $\pm$ 0.1} & \textbf{1.1 $\pm$ 0.1} & \textbf{1.1 $\pm$ 0.1} & 0.6 $\pm$ 0.2 \\
        \hline
        Average       & -17.54 & -12.87 & -9.38 & -6.41 & -9.63 & -12.90  \\
    \end{tabular}
    \caption{Mean DCS ($\pm$ S.E.) using summed probabilities over 10 abstention phrase variations across tested benchmarks and models. All scores are multiplied by $100$ for readability. The best (highest) DCS per benchmark is shown in \textbf{bold} and the second best is \underline{underlined}.}\label{tab:DCS-multi-idk-results}
\end{table}

\newpage
\section*{NeurIPS Paper Checklist}

\begin{enumerate}

\item {\bf Claims}
    \item[] Question: Do the main claims made in the abstract and introduction accurately reflect the paper's contributions and scope?
    \item[] Answer: \answerYes{}
    \item[] Justification: The abstract and introduction describe the Distributional Correctness Score (DCS), theoretical properties, benchmark adaptations, and empirical findings on six models and twelve benchmarks, matching the theory and experiments in the main text.

\item {\bf Limitations}
    \item[] Question: Does the paper discuss the limitations of the work performed by the authors?
    \item[] Answer: \answerYes{}
    \item[] Justification: The paper includes an explicit ``Limitations'' section (Section~\ref{sec:limitations}) discussing the multiple-choice/explicit-abstention scope, sensitivity to a single canonical abstention phrase versus a multi-phrase IDK sum (with the multi-phrase appendix table as evidence), the restricted model set (six models up to 8B on consumer-grade GPUs), dependence on a pinned evaluation framework and scoring convention, and the small sample sizes of the follow-up optimisation and predictive-validity studies.

\item {\bf Theory assumptions and proofs}
    \item[] Question: For each theoretical result, does the paper provide the full set of assumptions and a complete (and correct) proof?
    \item[] Answer: \answerYes{}
    \item[] Justification: Definitions and propositions state setup (e.g., discrete answer sets, loadings), and formal proofs appear in the appendix alongside the main theoretical discussion.

    \item {\bf Experimental result reproducibility}
    \item[] Question: Does the paper fully disclose all the information needed to reproduce the main experimental results of the paper to the extent that it affects the main claims and/or conclusions of the paper (regardless of whether the code and data are provided or not)?
    \item[] Answer: \answerYes{}
    \item[] Justification: The experiments section specifies evaluated models, twelve public benchmarks, comparison metrics, use of Hugging Face Transformers, log-likelihood-based scoring, compute (single NVIDIA RTX~3060/3080, under one second per question on average), and points to an explicit license summary; appendices document prompting templates, benchmark and model background, and additional metric tables needed to replicate analyses.

\item {\bf Open access to data and code}
    \item[] Question: Does the paper provide open access to the data and code, with sufficient instructions to faithfully reproduce the main experimental results, as described in supplemental material?
    \item[] Answer: \answerYes{}
    \item[] Justification: The DCS metric and evaluation tasks are implemented in the public \texttt{eval-framework} library (linked in the Code and reproducibility statement). The benchmarks used are publicly available Hugging Face datasets.

\item {\bf Experimental setting/details}
    \item[] Question: Does the paper specify all the training and test details (e.g., data splits, hyperparameters, how they were chosen, type of optimizer) necessary to understand the results?
    \item[] Answer: \answerYes{}
    \item[] Justification: The study is inference-only scoring on established benchmarks; the paper specifies models, benchmark list, metric definitions, prompting, and length-normalised log-likelihoods, with standard splits implied by the cited evaluation framework and benchmark sources.

\item {\bf Experiment statistical significance}
    \item[] Question: Does the paper report error bars suitably and correctly defined or other appropriate information about the statistical significance of the experiments?
    \item[] Answer: \answerYes{}
    \item[] Justification: Tables report means with standard errors, figure captions state that error bars are standard errors, and the text reports unpaired $t$-tests with $p$-values and a $95\%$ confidence interval for a key MMLU comparison.

\item {\bf Experiments compute resources}
    \item[] Question: For each experiment, does the paper provide sufficient information on the computer resources (type of compute workers, memory, time of execution) needed to reproduce the experiments?
    \item[] Answer: \answerYes{}
    \item[] Justification: The experiments section reports a single-workstation NVIDIA GeForce RTX~3060 or RTX~3080 (one consumer GPU, no cluster), typical per-question GPU wall time under one second, and that initial pilots used the same hardware class with modest cost relative to final runs.
    
\item {\bf Code of ethics}
    \item[] Question: Does the research conducted in the paper conform, in every respect, with the NeurIPS Code of Ethics \url{https://neurips.cc/public/EthicsGuidelines}?
    \item[] Answer: \answerYes{}
    \item[] Justification: The work analyses public language models on standard published benchmarks using established inference tooling, with no human subjects component or deceptive practices.

\item {\bf Broader impacts}
    \item[] Question: Does the paper discuss both potential positive societal impacts and negative societal impacts of the work performed?
    \item[] Answer: \answerYes{}
    \item[] Justification: The paper motivates evaluation reform for hallucination and overconfidence (positive), and discusses risks of confident wrong answers versus abstention in safety-relevant settings such as TruthfulQA and Winogender.
    
\item {\bf Safeguards}
    \item[] Question: Does the paper describe safeguards that have been put in place for responsible release of data or models that have a high risk for misuse (e.g., pre-trained language models, image generators, or scraped datasets)?
    \item[] Answer: \answerNA{}
    \item[] Justification: The submission proposes an evaluation metric and analyses publicly available models and benchmarks; it does not release a new high-risk model, dataset scrape, or similar asset requiring additional release safeguards beyond standard use of existing public artifacts.

\item {\bf Licenses for existing assets}
    \item[] Question: Are the creators or original owners of assets (e.g., code, data, models), used in the paper, properly credited and are the license and terms of use explicitly mentioned and properly respected?
    \item[] Answer: \answerYes{}
    \item[] Justification: The appendix ``Licenses and terms of use'' tabulates primary license names for inference libraries, each evaluated checkpoint, and each benchmark corpus, with citations provided.

\item {\bf New assets}
    \item[] Question: Are new assets introduced in the paper well documented and is the documentation provided alongside the assets?
    \item[] Answer: \answerNA{}
    \item[] Justification: The contribution is a metric and empirical analysis using existing benchmarks and models; no new dataset, model checkpoint, or code package is released in the paper (although the code implementing the metric is released and available in the Apache 2.0 \textsc{eval-framework} package).

\item {\bf Crowdsourcing and research with human subjects}
    \item[] Question: For crowdsourcing experiments and research with human subjects, does the paper include the full text of instructions given to participants and screenshots, if applicable, as well as details about compensation (if any)? 
    \item[] Answer: \answerNA{}
    \item[] Justification: The paper does not conduct new crowdsourcing or human-subjects experiments.

\item {\bf Institutional review board (IRB) approvals or equivalent for research with human subjects}
    \item[] Question: Does the paper describe potential risks incurred by study participants, whether such risks were disclosed to the subjects, and whether Institutional Review Board (IRB) approvals (or an equivalent approval/review based on the requirements of your country or institution) were obtained?
    \item[] Answer: \answerNA{}
    \item[] Justification: No new human-subjects or crowdsourcing study is performed in this work.

\item {\bf Declaration of LLM usage}
    \item[] Question: Does the paper describe the usage of LLMs if it is an important, original, or non-standard component of the core methods in this research? Note that if the LLM is used only for writing, editing, or formatting purposes and does \emph{not} impact the core methodology, scientific rigor, or originality of the research, declaration is not required.
    \item[] Answer: \answerYes{}
    \item[] Justification: LLMs are evaluated, as described in the paper and appendix; they are not used to evaluate themselves, \textit{e.g.}, as judges.

\end{enumerate}

\end{document}